\PassOptionsToPackage{table,usenames,dvipsnames}{xcolor}
\documentclass[table]{gtech}
\usepackage{xcolor}
\usepackage{amssymb}
\usepackage{multirow}
\usepackage{bigdelim}
\usepackage{longtable}
\usepackage{tabularray}
\usepackage{wrapfig}
\usepackage[most]{tcolorbox}
\usepackage{url}
\usepackage{float}
\usepackage{enumitem}

\usepackage{subcaption}
\RequirePackage{tgpagella} 
\RequirePackage{mathpazo}  
\RequirePackage{inconsolata} 
\usepackage{makecell}

\usepackage{adjustbox}
\usepackage{tablefootnote}
\usepackage{threeparttable}

\usepackage{booktabs} 
\usepackage{array}    
\newcolumntype{C}[1]{>{\centering\arraybackslash}m{#1}}

\usepackage{pifont}
\newcommand{\cmark}{\ding{51}}
\newcommand{\xmark}{\ding{55}}

\usepackage[utf8]{inputenc} 
\usepackage[T1]{fontenc}    
\usepackage{hyperref}       
\usepackage{url}            
\usepackage{booktabs}       
\usepackage{amsfonts}       
\usepackage{nicefrac}       
\usepackage{microtype}      
\usepackage{xspace}
\usepackage{amsthm}   
\usepackage{amsmath}  
\usepackage{amssymb}  
\usepackage{bm}       
\usepackage{tabularx}
\usepackage{array}
\usepackage{arydshln}

\newcolumntype{Y}{>{\centering\arraybackslash}X}

\theoremstyle{definition}

\renewcommand{\title}[1]{\newcommand{\titlelist}{{\huge\fontfamily{optimistic}\selectfont #1}}}

\newcommand{\ignore}[1]{}

\usepackage{arydshln}
\definecolor{CQColor}{rgb}{0.0,0.0,1.0} 

\usepackage{graphicx}
\usepackage{colortbl}
\usepackage{amssymb}
\usepackage{pifont}
\usepackage{booktabs,multirow}
\usepackage{makecell}
\usepackage{tabulary}
\usepackage{fontawesome5}
\usepackage{bbding}
\usepackage{multicol}
\newcommand \blfootnote[1]{
    \begingroup
        \renewcommand
        \thefootnote{}\footnote{#1}
        \addtocounter{footnote}{-1}
        \vspace{-1ex}
    \endgroup
}
\newlength\savewidth

\providecommand{\benchmarkname}{\textsc{FinFIRST}}

\providecommand{\nummodels}{\textsc{15}}

\title{
\huge
\textcolor[HTML]{7FB3FF}{Fin}\textcolor[HTML]{0369ff}{FIRST:}
Benchmarking Search Agents for
\textcolor[HTML]{0369ff}{F}inancial
\textcolor[HTML]{0369ff}{I}nformation
\textcolor[HTML]{0369ff}{R}etrieval,
\textcolor[HTML]{0369ff}{S}ourcing and
\textcolor[HTML]{0369ff}{T}raceability
}

\author[1,*]{Wenqing Wang}
\author[1,*]{Haitao Xiang}
\author[1,2,\diamond]{Xinyi Zhao}
\author[1]{Mingming Yin}
\author[1]{Ying Zhong}
\author[1]{Zhaoxin Huan}
\author[3]{Qiheng Zhou}
\author[1]{Jin Zhu}
\author[1,\dagger]{Xiaolu Zhang}
\author[1]{Shi Chang}
\author[1]{Jun Zhou}

\contribution[1]{Ling Team, Inclusion AI}

\contribution[2]{Peking University}

\contribution[3]{China International Capital Corporation Limited}

\abstract{

Financial search is a highly demanding task for LLM agents, requiring not only a correct final answer but also temporally valid information retrieval, authoritative source selection, entity and period alignment, unit and definition consistency, and verifiable evidence for all conclusions. Existing benchmarks predominantly evaluate only the final answer, making it difficult to localize errors or assess whether an answer is well-founded. To address this gap, we introduce \textbf{\benchmarkname{}} (\textbf{\textsc{F}}inancial \textbf{\textsc{I}}nformation \textbf{\textsc{R}}etrieval, \textbf{\textsc{S}}ourcing and \textbf{\textsc{T}}raceability), the first financial benchmark to jointly evaluate answers and supporting evidence through atomic rubrics. \benchmarkname{} comprises 123 expert-authored tasks spanning a graduated difficulty spectrum, constructed from aggregate patterns of real-world financial scenarios via an 18-field taxonomy, a 6-axis coverage blueprint, a registry of 138 financial sources, contributions from over 50 finance experts, and 6-stage quality control pipeline. Each task is accompanied by an evidence-grounded reference package decomposed into atomic criteria across three dimensions—raw-information acquisition, source verification, and computation and answer formation—enabling fine-grained evaluation of the complete research process. We evaluate \nummodels{} model configurations under a unified tool setting: Claude-Opus-5 achieves the highest atomic score (87.59\%), while GPT-5.6-Sol attains the highest strict pass rate (71.54\%), with computation and answer formation consistently lagging behind raw-information acquisition as a common bottleneck across all systems. \benchmarkname{} thus retains final-answer correctness as the primary objective while making the supporting research process measurable, verifiable, and diagnosable \footnote{This paper presents \benchmarkname{} V1; the forthcoming V2 will expand the benchmark further, with tasks 

spanning foundational search and advanced financial research.}.
}

\date{September, 2026}
\gtechdata[Huggingface]{\url{https://huggingface.co/datasets/inclusionAI/FinFIRST}}

\begin{document}
\blfootnote{%
$^{*}$Equal contribution.\quad
$^{\dagger}$Corresponding author:  <\email{yueyin.zxl@antgroup.com}>.\quad
$^{\diamond}$Work done during an internship at Ling Team, Inclusion AI.
}

\maketitle
\section{Introduction}
\label{sec:intro}

Search has emerged as a core capability of Large Language Model (LLM) agents \citep{browsecomp,mind2web2}, and finance provides a particularly demanding setting for this capability \citep{finsearchcomp,financeagent}. A successful search in finance rarely means locating a single webpage or matching a number.
Instead, an agent must identify authoritative and time-valid sources, retrieve information for the correct entity and reporting period, align units and financial definitions, reconcile evidence across sources, perform necessary calculations, and support its conclusion with verifiable evidence. Errors at any of these steps can yield a plausible-looking but unsupported answer.
Evaluating financial research agents therefore requires examining not only the final answer, but also the evidence and intermediate steps used to derive it.

However, existing benchmarks cover complementary parts of this workflow. Financial QA datasets such as FinQA \citep{finqa}, ConvFinQA \citep{convfinqa}, TAT-QA \citep{tatqa}, and FinanceBench \citep{financebench} provide the relevant evidence context, and therefore primarily evaluate reasoning over evidence rather than the end-to-end process that discovers it.

Finance Agent Benchmark \citep{financeagent} and FrontierFinance \citep{frontierfinance} extend evaluation to tool use and research-style outputs, but do not directly assess supporting evidence through atomic intermediate criteria. FinDeepIndicator \citep{findeepindicator} evaluates intermediate stages, but focuses on template-derived financial indicator construction. 
Among these agentic financial benchmarks, FinSearchComp \citep{finsearchcomp} is most closely related to our work: it comprises 635 expert-curated questions covering time-sensitive data retrieval, historical lookup, and complex investigation across global and Greater China markets. 
Nevertheless, its binary final-answer evaluation reveals neither where an error occurs---in source selection, data extraction, definition alignment, or calculation---nor whether a correct response is genuinely supported by appropriate evidence or merely happens to match the reference answer. Table~\ref{tab:benchmark-comparison} summarizes these differences.

\begin{table*}[t]
\centering
\caption{
Comparison of representative financial benchmarks.
\textbf{Agentic}: requires active information retrieval or external tool use.
\textbf{Time-sensitive}: correct answers are tied to a specific time points and may change as new information emerges. 
\textbf{Demand-aligned}: reflects authentic financial needs or real-world professional workflows.
\textbf{Expert-authored}: designed and verified by domain experts in finance.
\textbf{Multi-source}: integrates and reconciles evidence from multiple heterogeneous sources. 
\textbf{Evidence scoring}: assesses the quality and relevance of supporting sources and evidence. 
\textbf{Atomic assessment}: decomposes tasks into rubric-specific criteria rather than evaluating final answers correctness.
}
\label{tab:benchmark-comparison}
\scriptsize
\setlength{\tabcolsep}{3.3pt}
\resizebox{\textwidth}{!}{%
\begin{tabular}{@{}lccccccc@{}}
\toprule
Benchmark &
\textbf{Agentic} &
\begin{tabular}[c]{@{}c@{}}
\textbf{Time-}\\\textbf{sensitive}
\end{tabular} &
\begin{tabular}[c]{@{}c@{}}
\textbf{Demand-}\\\textbf{aligned}
\end{tabular} &
\begin{tabular}[c]{@{}c@{}}
\textbf{Expert-}\\\textbf{authored}
\end{tabular} &
\begin{tabular}[c]{@{}c@{}}
\textbf{Multi-}\\\textbf{source}
\end{tabular} &
\begin{tabular}[c]{@{}c@{}}
\textbf{Evidence-}\\\textbf{scoring}
\end{tabular} &
\begin{tabular}[c]{@{}c@{}}
\textbf{Atomic-}\\\textbf{assessment}
\end{tabular} \\
\midrule

ConvFinQA~\citep{convfinqa}
& \xmark & \xmark & \xmark & \textcolor[HTML]{0369ff}\cmark
& \xmark & \xmark & \xmark \\

FinanceBench~\citep{financebench}
& \xmark & \xmark & \xmark & \textcolor[HTML]{0369ff}\cmark
& \xmark & \xmark & \xmark \\

FinanceQA~\citep{financeqa}
& \xmark & \xmark & \xmark & \textcolor[HTML]{0369ff}\cmark
& \xmark & \xmark & \xmark \\

BizFinBench~\citep{bizfinbench}
& \xmark & \xmark & \textcolor[HTML]{0369ff}\cmark & \textcolor[HTML]{0369ff}\cmark
& \xmark & \xmark & \xmark \\

Finance Agent Benchmark~\citep{financeagent}
& \textcolor[HTML]{0369ff}\cmark & \xmark & \xmark & \textcolor[HTML]{0369ff}\cmark
& \textcolor[HTML]{0369ff}\cmark & \xmark & \textcolor[HTML]{0369ff}\cmark \\

FinSearchComp~\citep{finsearchcomp}
& \textcolor[HTML]{0369ff}\cmark & \textcolor[HTML]{0369ff}\cmark & \xmark & \textcolor[HTML]{0369ff}\cmark
& \textcolor[HTML]{0369ff}\cmark & \xmark & \xmark \\

FrontierFinance (research)~\citep{frontierfinance}
& \textcolor[HTML]{0369ff}\cmark & \textcolor[HTML]{0369ff}\cmark & \xmark & \textcolor[HTML]{0369ff}\cmark
& \textcolor[HTML]{0369ff}\cmark & \textcolor[HTML]{0369ff}\cmark & \textcolor[HTML]{0369ff}\cmark \\

FrontierFinance (computer use)~\citep{frontierfinancecomputer}
& \textcolor[HTML]{0369ff}\cmark & \xmark & \xmark & \textcolor[HTML]{0369ff}\cmark
& \textcolor[HTML]{0369ff}\cmark & \xmark & \textcolor[HTML]{0369ff}\cmark \\

FinDeepIndicator~\citep{findeepindicator}
& \textcolor[HTML]{0369ff}\cmark & \xmark & \xmark & \xmark
& \textcolor[HTML]{0369ff}\cmark & \xmark & \textcolor[HTML]{0369ff}\cmark \\
\noalign{\vskip 0.08cm}
\hdashline
\noalign{\vskip 0.08cm}
\textbf{\benchmarkname{} (ours)}
& \textcolor[HTML]{0369ff}\cmark & \textcolor[HTML]{0369ff}\cmark & \textcolor[HTML]{0369ff}\cmark & \textcolor[HTML]{0369ff}\cmark
& \textcolor[HTML]{0369ff}\cmark & \textcolor[HTML]{0369ff}\cmark & \textcolor[HTML]{0369ff}\cmark \\

\bottomrule 
\end{tabular}%
}
\end{table*}

\begin{table*}[t]
\centering
\caption{
An example combining Meta's official disclosures with an industry report, and its atomic evaluation rubric.
}
\label{tab:finfirst-example}
\small
\setlength{\tabcolsep}{5pt}
\renewcommand{\arraystretch}{1.12}

\begin{tabularx}{\textwidth}{
    @{}
    >{\raggedright\arraybackslash}p{0.18\textwidth}
    >{\raggedright\arraybackslash}X
    >{\centering\arraybackslash}p{0.07\textwidth}
    @{}
}
\hline
\toprule
\multicolumn{3}{@{}p{0.97\textwidth}@{}}{
\textbf{\textcolor[HTML]{0369ff}{Question.}}
Using the latest official and industry data available as of June 30, 2026, and Meta's official disclosures together with the IAB/PwC Internet Advertising Revenue Report, calculate Meta's 2025 United States \& Canada advertising revenue as a percentage of 2025 U.S. social media advertising revenue. Report the result as a percentage rounded to two decimal places.
} \\[3pt]

\multicolumn{3}{@{}p{0.97\textwidth}@{}}{
\textbf{\textcolor[HTML]{0369ff}{Reference answer.}}
Meta reported quarterly U.S.\ \& Canada advertising revenue of
\$18.259, \$20.045, \$21.331, and \$25.643 billion, totaling
\$85.278 billion. The IAB/PwC report gives 2025 U.S.\ social media
advertising revenue of \$117.70 billion. Therefore,
$\$85.278/\$117.70 \times 100\% = \mathbf{72.45\%}$.
} \\
\midrule

\textbf{\textcolor[HTML]{0369ff}{Capability}} & \textbf{\textcolor[HTML]{0369ff}{Atomic criterion}} & \textbf{\textcolor[HTML]{0369ff}{Points}} \\
\noalign{\vskip 0.08cm}
\hdashline
\noalign{\vskip 0.08cm}

\textit{Source verification}
& Identifies Meta's official 2025 disclosure materials or filing.
& 10 \\

\textit{Source verification}
& Identifies the 2025 IAB/PwC Internet Advertising Revenue Report.
& 10 \\

\textit{Raw-information acquisition}
& Extracts Meta's four quarterly U.S.\ \& Canada advertising-revenue figures: \$18.259, \$20.045, \$21.331, and \$25.643 billion.
& 30 \\

\textit{Raw-information acquisition}
& Extracts 2025 U.S.\ social media advertising revenue of \$117.70 billion.
& 20 \\

\textit{Computation and answer formation}
& Correctly aggregates Meta's quarterly revenue:
$18.259+20.045+21.331+25.643=85.278$ billion.
& 10 \\

\textit{Computation and answer formation}
& Correctly calculates:
$85.278/117.70 \times 100\%=72.45\%$.
& 10 \\

\textit{Computation and answer formation}
& Reports the final answer as 72.45\%, with the required precision and unit.
& 10 \\

\noalign{\vskip 0.08cm}
\hdashline
\noalign{\vskip 0.08cm}
\textbf{Total} & & \textbf{100} \\
\hline
\toprule
\end{tabularx}
\end{table*}

In this work, we introduce \textbf{\benchmarkname{}}
(Financial Information Retrieval, Sourcing and
Traceability), the first financial benchmark to evaluate both generated answers and retrieved evidence through atomically decomposed rubrics.
The rubrics span three capability groups: \textit{(1) Raw-information acquisition} examines whether the required facts and figures are retrieved with the correct entity, period, unit, and definition;
\textit{(2) Source verification} assesses whether the cited sources are authoritative, temporally valid, and preferably first-party; and \textit{(3) Computation and answer formation} evaluates subsequent calculations, reasoning, completeness, and instruction following. Traceability is enforced throughout by requiring key facts, figures, intermediate calculations, and conclusions to be explicitly grounded in and traceable to the cited evidence.  The construction of \benchmarkname{} benefited from professional support provided by the investment banking team at China International Capital Corporation Limited (CICC), lending strong industry-grounded expertise to the development process.

 Specifically, \benchmarkname{} is constructed through the three-phase pipeline.
First, we derive an 18-field taxonomy and a 6-axis coverage blueprint from aggregate patterns in real financial search demand.
In parallel, finance experts compile a routing map of 138 sources, annotating each for authority, task suitability, and geographic coverage.
Second, over 50 finance experts author questions guided by the coverage blueprint, with continuous rebalancing to fill under-represented categories. Guided by the source map, they construct a reference package for each question, including the answer, supporting sources, key definitions and data versions, calculations, valid alternatives, and numerical tolerances. 
 Each package is then decomposed into atomic rubric items across 
 \textit{raw-information acquisition}, \textit{source verification}, and \textit{computation and answer formation}. 
Third, every candidate undergoes 6-stage quality control: scope review, independent re-solving, cross-validation and adjudication, rubric auditing, model-based stress testing, and final consistency checks.
This pipeline yields 123 realistic, diverse, and expert-vetted tasks with graduated difficulty, ranging from focused single-source lookups to complex multi-step research requiring planning, cross-source synthesis, and calculation.
Table~\ref{tab:finfirst-example} presents an advanced-research example alongside its atomic rubric.

We evaluate \nummodels{} model configurations under a unified toolset comprising web search, webpage fetching, and Python execution, employing atomic scores and strict pass rates to distinguish partial progress from complete task success. 
The results yield three main findings. First, even the strongest agents remain far from perfect: Claude-Opus-5 achieves the highest overall rubric score (87.59\%), while GPT-5.6-Sol attains the highest strict pass rate (71.54\%). The gap between these two metrics suggests that high aggregate scores often reflect partially correct research processes rather than consistently complete solutions. Second, \textit{computation and answer formation} consistently lag behind \textit{raw-information acquisition} across all evaluated systems, indicating that successfully retrieving relevant data does not guarantee correct calculation, synthesis, or reporting. Third, similar overall scores can mask distinct capability profiles. For instance, GLM-5.3 and Kimi-K3 achieve nearly identical overall scores (80.61\% vs. 80.83\%), yet GLM-5.3 substantially outperforms Kimi-K3 in source verification (78.56\% vs. 70.70\%), whereas Kimi-K3 holds an advantage in raw-information acquisition and computation and answer formation. Together, these findings demonstrate the diagnostic value of atomic assessment in pinpointing where financial research agents succeed and where they fall short.

Our contributions are threefold:
\begin{itemize}
    \item We introduce and release \benchmarkname{}, a demand-aligned, expert-authored benchmark of 123 tasks with graduated difficulty, accompanied by atomic rubrics that evaluate both answers and evidence for fine-grained assessment.
    
    \item We develop an expert-led construction pipeline grounded in real-world financial scenarios,
    combining an 18-field taxonomy, a 6-axis coverage blueprint, 138 financial sources, more than 50 finance experts, and 6-stage quality control.
    
    \item We systematically evaluate \nummodels{} model configurations under a unified basic-tool setting, revealing a 
    gap between atomic scores and strict task completion, a consistent bottleneck in computation and answer formation, and distinct capability profiles hidden by similar aggregate scores.
\end{itemize}


\section{Data Construction}
\begin{figure*}[ht]
    \centering
    \includegraphics[width=\linewidth]{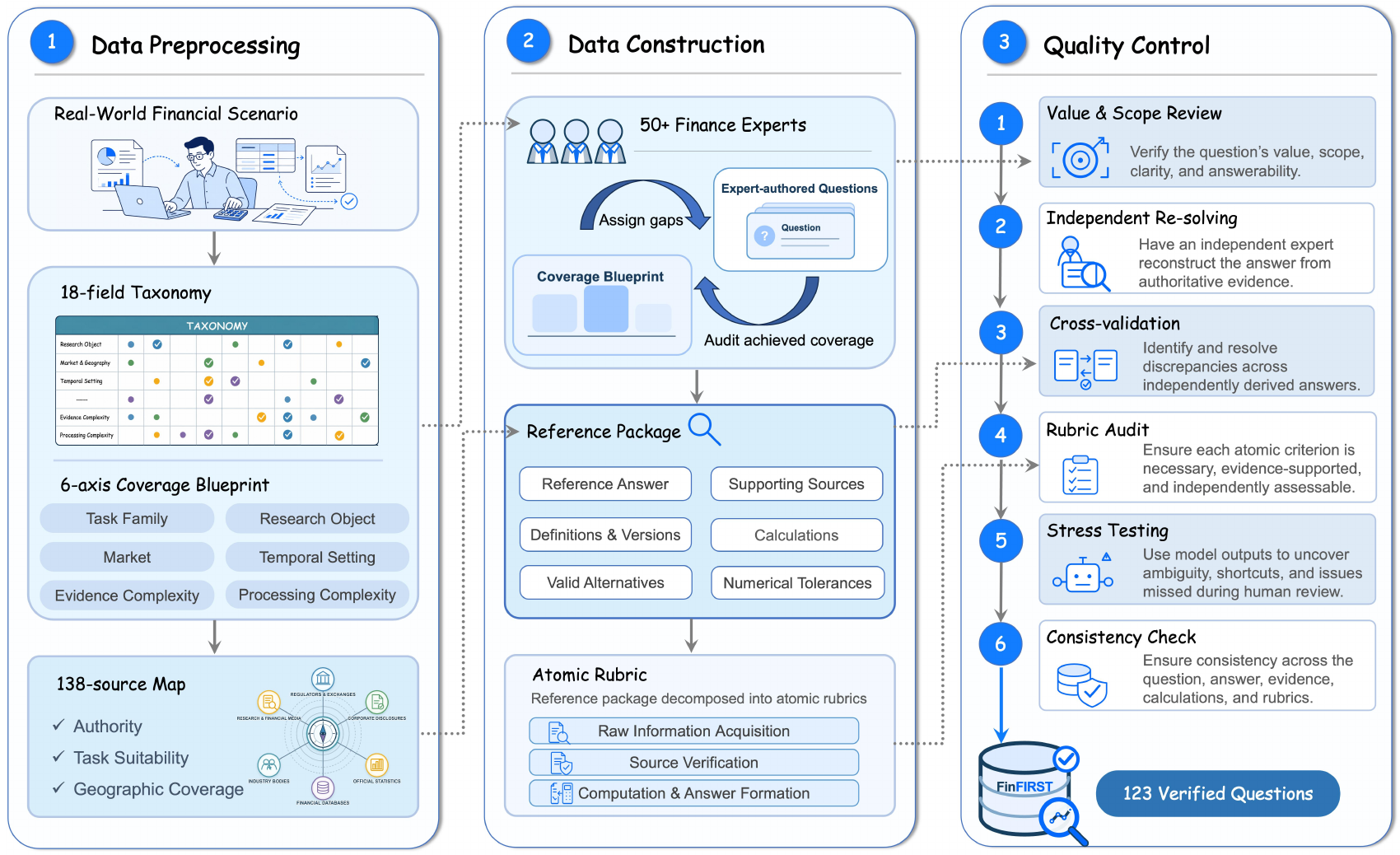}
    \caption{Overview of \benchmarkname{} construction pipeline. The pipeline comprises three stages: data preprocessing derives an 18-field taxonomy, a 6-axis coverage blueprint, and a 138-source map from patterns observed in real-world financial scenarios;
    data construction engages over 50 finance experts to iteratively author questions, balance coverage, construct evidence-grounded reference packages, and define atomic rubrics; and quality control applies a 6-stage review and filtering procedure, resulting in a final set of 123 verified questions.    
    }
    \label{fig:finfirst-construction-overview}
    \end{figure*}

The construction of \benchmarkname{} follows four principles: 
benchmark composition should reflect financial information needs observed in practice; task difficulty should arise from meaningful retrieval, source selection, definition alignment, evidence reconciliation, and computation rather than obscure facts or artificial traps; each reference answer should be reproducible from authoritative, version-correct evidence; and evaluation should reveal where an agent succeeds or fails within the research process.
As illustrated in Figure~\ref{fig:finfirst-construction-overview}, we operationalize these principles through three phases: data preprocessing, expert-led data construction, and six-stage quality control.

\subsection{Data Preprocessing}
\label{sec:data-preprocessing}

\subsubsection{Demand Analysis and Taxonomy Design}
We abstract recurring financial information needs into an 18-field taxonomy, organized across four semantic groups.
\textbf{Task structure} captures question decomposition, subquestion dependencies, and conditional filtering. \textbf{Research content} covers the research object, target metric, metric definition or accounting convention, geography or market, temporal-constraint type, and explicit time specification. \textbf{Processing requirements} specify the verification mode, computation type, output format, and response language. \textbf{Source and retrieval intent} records whether a source is specified, its accessibility and mainstreamness, the number of sources required, and the retrieval intent—data retrieval, fact retrieval, evidence localization, or concept identification. 
Notably, these four groups constitute an organizational summary rather than a rigid four-tuple: every question is annotated over a uniform 18-field schema, with individual fields may be multi-valued or inapplicable. 
This representation supports stratified sampling, coverage auditing, and fine-grained capability analysis.
Aggregate patterns from real-world financial-use scenarios determine the baseline distribution, while professionally important but less frequent settings are retained when they can be supported by accessible and reproducible evidence.

\subsubsection{Coverage Blueprint}

Taking the Cartesian product of all 18 taxonomy fields would yield a large number of implausible or redundant tasks. We therefore define a coverage blueprint over six dimensions that directly shape the research process: \textbf{task structure}, \textbf{research object}, \textbf{market}, \textbf{temporal setting}, \textbf{evidence complexity}, and \textbf{processing complexity}. Evidence complexity distinguishes specified-source retrieval from open search and single-source tasks from multi-source research; processing complexity ranges from direct extraction and conditional filtering to financial calculation and multi-step synthesis. Together, these two dimensions jointly determine the difficulty distribution: easy questions involve direct or limited-step retrieval, medium questions require additional alignment, verification, or straightforward calculation, and hard questions combine multiple research steps, sources, definitions, periods, calculations, or synthesis requirements.

\subsubsection{Financial Source Map}

In parallel, finance experts construct a routing map of 138 financial sources spanning major markets and research objects. Each source is annotated with its \textbf{identity}, \textbf{access path}, \textbf{description}, \textbf{authority level}, \textbf{applicable entities and metrics}, \textbf{task suitability}, and \textbf{geographic coverage}. The map covers regulators and exchanges, statutory and corporate disclosures, official statistics, international and industry organizations, professional financial databases, specialist research providers, and established financial media. Its authority hierarchy prioritizes statutory, official, and first-party evidence, while allowing established secondary sources when they are appropriate to the task and can be corroborated.

The source map is a routing aid rather than an answer key. For each question, experts jointly consider the research object, metric, market, temporal requirement, and verification objective to identify suitable primary sources and valid alternatives. The exact documents and evidence accepted for an individual question are recorded separately in its reference package.




\subsection{Expert-Led Data Construction}
\label{sec:expert-construction}

\subsubsection{Question Authoring}


More than 50 finance professionals participate in question construction. Experts author new questions according to the taxonomy and coverage blueprint, with each question assigned structured labels under the same schema used during demand analysis. Coverage is monitored continuously by comparing the distribution of candidate questions against the target; experts are directed to fill documented gaps while redundant or over-represented candidates are revised or removed, creating an iterative loop between authoring and coverage auditing.


Each question must define the research target with sufficient precision. Where relevant to the answer, the prompt specifies the target entity, temporal range or cutoff, financial or statistical definition, unit, expected output, and numerical precision. Questions whose difficulty derives primarily from ambiguity, inaccessible evidence, obscure trivia, or unnecessary computation are revised or excluded.




\subsubsection{Reference-Package Construction}

For each candidate question, the expert uses the source map to identify task-appropriate primary sources and valid alternatives, then constructs an item-specific reference package. The package contains: reference answer; authoritative documents or URLs with precise evidence locations, such as the filing, section, page, table, or data-series identifier; the required facts and numerical inputs; relevant definitions, reporting periods, units, currencies, consolidation scopes, and data versions; intermediate calculations or synthesis steps; and acceptable alternative sources, equivalent definitions, and numerical tolerances. For time-sensitive items, the package additionally records the reference cutoff, publication or filing date, access date, and evidence version, collectively preserving reproducibility when the underlying information is subsequently updated, revised, or restated.





\subsubsection{Atomic Rubric Construction}
\label{sec:atomic-rubrics}
Finance experts further decompose the reference package into independently assessable atomic criteria, each of which must be required by the question, supported by the reference package, and specific enough to evaluate without unstated assumptions. The criteria span three capability groups: \textbf{raw-information acquisition}, covering whether each required fact or value is retrieved with the correct entity, period, unit, currency, and definition; \textbf{source verification}, assessing whether the agent uses authoritative, task-relevant, and time- and version-valid evidence; and \textbf{computation and answer formation}, measuring the correctness of the required calculations or reasoning and the completeness and format of the final answer.
Traceability is enforced across all three groups by linking key facts, numerical inputs, intermediate calculations, and conclusions to supporting evidence. When financially equivalent solutions exist, the rubric records valid alternative evidence paths, equivalent definitions, and explicit numerical tolerances, thereby rewarding correct intermediate work, avoiding the treatment of a single reference path as the only valid solution, and preventing unsupported answers from receiving full credit merely by matching the final value.






\subsection{Quality Control}
\label{sec:quality-control}


Every candidate undergoes the same six-stage quality-control process; those that fail a stage are revised and reassessed or removed.

\begin{itemize}
    \item \textbf{Value and scope review.}
    Reviewers verify that the question represents a meaningful financial need with a clear entity, period, definition, source requirement, and answer boundary.
    Questions based on artificial search traps, immaterial facts, or complexity unrelated to financial research are revised or rejected.


    
    \item \textbf{Independent re-solving.} A second finance expert solves the question from scratch, independently locating authoritative evidence, verifying document versions and reporting periods, extracting the required inputs, and reproducing the calculation or analysis.


    \item \textbf{Cross-validation and adjudication.} The independent solution from the second finance expert is compared with the original reference package, and any disagreements are localized to their source, version, entity scope, definition, period, unit, currency, calculation, or interpretation and resolved against the strongest available evidence rather than by majority vote.

    \item \textbf{Rubric audit.}
    Reviewers verify that each atomic criterion is explicitly required by the question, supported by the reference evidence, assigned to the appropriate capability group, and weighted consistently.
    The rubric must not introduce source, format, or intermediate-result requirements that are absent from the prompt.
    Any revision to the question, answer, evidence, calculation, or tolerance is propagated to the other components.

    \item \textbf{Model-based stress testing.}
    Candidate questions are executed by multiple LLMs, including \textsc{Claude-Opus-5}~\citep{claudeopus5}, \textsc{GPT-5.6-Sol}~\citep{gpt56}, and \textsc{GLM-5.3}~\citep{glm53}, to expose residual ambiguity, outdated evidence, omitted sources, brittle scoring conditions, and unintended shortcuts. Model outputs are used only to identify potential issues and calibrate task difficulty; all revisions and acceptance decisions remain under expert control.

    \item \textbf{Final consistency check.}
    Accepted candidates undergo deduplication, leakage checks, taxonomy normalization, evidence-link and version verification, and cross-component consistency checks.
    Reviewers confirm that the prompt, reference package, atomic rubric, and taxonomy labels describe the same research task and evidence requirements.
\end{itemize}

\subsection{Data Statistics}
\label{sec:final-benchmark}

\begin{figure*}[t]
    \centering
    \includegraphics[width=\linewidth]{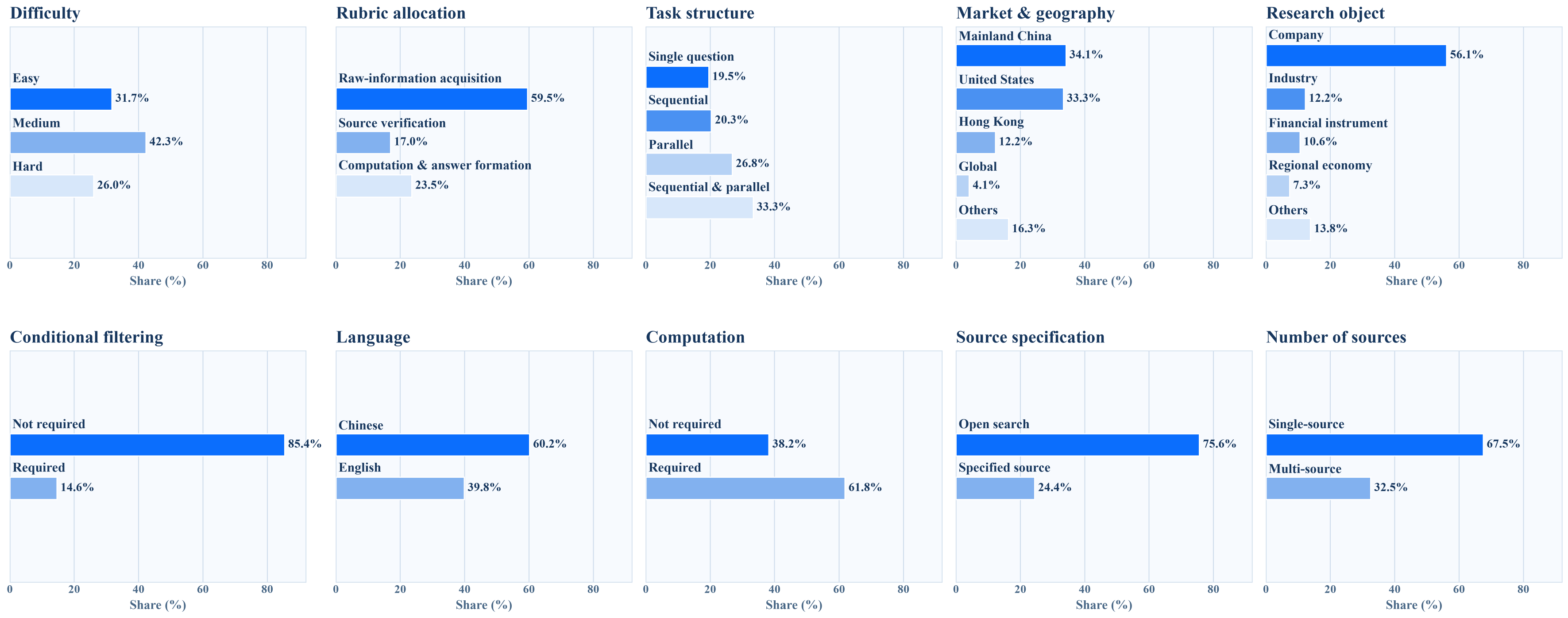}
    \caption{
    Composition of \benchmarkname{} across dataset and rubric dimensions.
    Each panel reports category shares within the corresponding dimension.
    Percentages are rounded, and
    long-tail categories are grouped under \emph{Others} where applicable.
    }
    \label{fig:data-composition}
\end{figure*}

After quality-control filtering, \benchmarkname{} retains 123 expert-authored and verified questions, corresponding to an overall item acceptance rate of 9.78\%. Figure~\ref{fig:data-composition} summarizes the benchmark along ten dimensions: difficulty, rubric allocation, task structure, research object, market and geography, language, computation, conditional filtering, and source specification and count.



The benchmark follows a graded difficulty distribution of 31.7\% easy, 42.3\% medium, and 26.0\% hard questions, with prompts and reference answers averaging 187.02 and 90.93 characters per task, respectively. Rubric points are distributed across raw-information acquisition (59.5\%), source verification (17.0\%) and computation and answer formation (23.5\%), reflecting the benchmark's dual emphasis on retrieving required information and producing evidence-supported answers.
The final composition combines broad coverage with substantial process complexity: more than four-fifths of the questions contain multiple related subproblems, 61.8\% require explicit computation, 32.5\% draw on multiple sources, and 14.6\% involve conditional filtering. While 75.6\% of questions require open search without a specified source, the remainder involve designated sources—all of which are publicly accessible. Companies constitute the largest research-object category, with mainland China and the United States contributing comparable shares of geographic coverage. The benchmark is bilingual, comprising 60.2\% Chinese and 39.8\% English questions.

\section{Experiment}

\subsection{Evaluation Setup}

We evaluate 15 model configurations on all 123 questions in \benchmarkname{} under a common agent harness. All models follow the same ReAct-style \citep{yao2023reactsynergizingreasoningacting} interaction protocol and share an identical toolset—web search, page visit, and Python execution—thereby reducing variation introduced by proprietary search stacks and keeping the focus on each agent's ability to plan searches, retrieve evidence, and construct well-supported answers. We set the output temperature $T$ of 1.0, and results were reported over one designated run for each model configuration. 
Appendix \ref{app:tool-environment} details the tool interfaces and their roles in the research workflow.


\subsection{Models}

A total of 15 model configurations are evaluated: \textsc{Qwen3.8-Max}~\citep{qwen38max}, \textsc{Qwen3.8-27B}~\citep{qwen3827b}, \textsc{Qwen3.8-Flash}~\citep{qwen38flash}, \textsc{MiniMax-M3}~\citep{minimaxm3}, \textsc{Hunyuan3-Thinking}~\citep{hunyuan3}, \textsc{GLM-5.3-Flash}~\citep{glm53flash}, \textsc{Gemini-3.7-Flash}~\citep{gemini37flash}, \textsc{GLM-5.2}~\citep{glm52}, \textsc{Claude-Opus-5}~\citep{claudeopus5}, \textsc{Kimi-K3}~\citep{kimik3}, \textsc{GLM-5.3}~\citep{glm53}, \textsc{DeepSeek-V4-Pro}~\citep{deepseekv4}, \textsc{DeepSeek-V4-Flash}~\citep{deepseekv4flash0731}, and \textsc{GPT-5.6-Sol}~\citep{gpt56}---together with the finance-specialized \textsc{Ling-3.0-Flash-Fin}~\citep{ling30flash-fin}. All results are produced under the common basic-tool setting described above. Detailed model cards and evaluated identifiers are provided in Appendix~\ref{app:model-cards}.

\subsection{Evaluation Metrics}

Each question $q$ contains a set of atomic rubric items $\mathcal{R}_q$, with the total count across all 123 questions summing to $\sum_{q\in\mathcal{Q}} |\mathcal{R}_q| = 701$. Each rubric item $i$ carries an expert-defined weight $w_{q,i}$, normalized within each question to 100 points ($\sum_{i\in\mathcal{R}_q} w_{q,i}=100$), yielding $123\times100=12{,}300$ weighted rubric points in total. Given a binary judge decision $z_{q,i}\in\{0,1\}$ for each item, we evaluate models using three complementary metrics:

\begin{equation}
\mathrm{Atomic} =
\frac{\sum_q\sum_{i\in\mathcal{R}_q}z_{q,i}}
{\sum_q |\mathcal{R}_q|},
\end{equation}

\begin{equation}
\mathrm{Weighted\ Rubric\ Score} =
\frac{\sum_q\sum_{i\in\mathcal{R}_q}w_{q,i}z_{q,i}}
{\sum_q\sum_{i\in\mathcal{R}_q}w_{q,i}},
\end{equation}

\begin{equation}
\mathrm{Strict\ Pass} =
\frac{1}{|\mathcal{Q}|}\sum_q
\mathbb{I}\!\left[\forall i\in\mathcal{R}_q,\ z_{q,i}=1\right].
\end{equation}

\noindent\textbf{Atomic} is the unweighted pass rate over all 701 criteria, treating every item equally regardless of its assigned weight. \textbf{Weighted Rubric Score} (referred to as \textbf{Loose} on the internal leaderboard) grants partial credit proportional to expert-assigned weights across the full 12,300-point benchmark. \textbf{Strict Pass} counts a question as correct only if all of its atomic criteria are satisfied.

Each atomic criterion is assigned to exactly one of three capability groups---raw-information acquisition, source verification, and computation and answer formation---with group-level scores reported as weighted pass rates over their respective point totals
\begin{equation}
W_g = \sum_{q\in\mathcal{Q}}
\sum_{i\in\mathcal{R}_{q,g}} w_{q,i},
\end{equation}
where $\mathcal{R}_{q,g}\subseteq\mathcal{R}_q$ denotes the criteria of question $q$ in group $g$. This yields 7,322 points for raw-information acquisition, 2,085 for source verification, and 2,893 for computation and answer formation, which together sum to the full benchmark total of 12,300.

\subsection{Rubric Judge}

We utilize GLM-5.1\citep{glm5} as the rubric judge. Given the question, the atomic rubric items, and the model's final response, the judge independently determines whether each item is fully satisfied—verifying numerical values, units, time ranges, financial definitions, sources, calculations, and output requirements, and rejecting any item with missing evidence, incorrect periods, or mismatched definitions. Scores are then computed deterministically from the resulting binary decisions and its weights. To validate judge reliability, we randomly sampled 50 evaluation instances for annotation by eight finance experts; after adjudication, item-level agreement between \textsc{GLM-5.1} and the human reference labels yielded Cohen’s Kappa $\kappa$ \citep{cohen1960coefficient} = 0.816, indicating strong agreement. The full judge prompt is provided in Appendix \ref{app:judge-prompt}, Table \ref{tab:judge-prompt}. 



\section{Result and Analysis}

\subsection{Overall Performance}


Table~\ref{tab:overall-results} reports performance and tool-use statistics for all 15 model configurations. \textsc{Claude-Opus-5} achieves the highest Atomic (87.59\%) and Loose Pass (87.61\%) scores, while \textsc{GPT-5.6-Sol} leads in Strict Pass rate with 88 of 123 questions fully satisfied (71.54\%). Even for the strongest systems, the gap between partial credit and complete task success remains substantial: \textsc{Claude-Opus-5}'s Strict Pass rate trails its Loose Pass score by 18.50 percentage points, 
indicating that a largely correct solution frequently retains at least one unresolved requirement.

The three rubric groups reveal a consistent downstream bottleneck: computation and answer formation underperforms raw-information acquisition for each model; the gap ranges from 5.27 percentage points for \textsc{GPT-5.6-Sol} to 21.80 points for \textsc{DeepSeek-V4-Flash}. 
Rubric decomposition further distinguishes models with similar aggregate performance: \textsc{GLM-5.3} and \textsc{Kimi-K3} achieve comparable Loose Pass scores (80.61\% and 80.83\%, respectively), yet \textsc{GLM-5.3} is stronger in source verification (78.56\% vs.\ 70.70\%), while \textsc{Kimi-K3} leads in both raw-information acquisition (84.84\% vs.\ 83.11\%) and computation and answer formation (77.98\% vs.\ 75.77\%). Among all evaluated systems, \textsc{Ling-3.0-Flash-Fin} stands out in source verification, reaching 82.45\%---the highest among open-weight models in the lower block of Table~\ref{tab:overall-results}.


Tool-use volume is not monotonically related to performance. \textsc{GPT-5.6-Sol} averages 62.28 tool calls per covered question versus 23.89 for \textsc{Claude-Opus-5}, and \textsc{Kimi-K3} records the largest single-question maximum at 313 calls; \textsc{Ling-3.0-Flash-Fin}, by contrast, pairs its strong source-verification score with a moderate average of 30.87 calls and a maximum of 119. Call counts should therefore be treated as efficiency indicators only when comparing systems at similar levels of task correctness, as higher volumes may reflect broad exploration, recovery from retrieval failures, or inefficient search rather than superior research outcomes.

\begin{table*}[t]
\centering
\caption{
Overall, rubric-level, and tool-efficiency results on \benchmarkname{} under the common basic-tool setting. $\downarrow$ indicates that lower values denote greater efficiency conditional on comparable task correctness, and efficiency statistics are averaged over questions with recorded trajectories. $^{\dagger}$ indicates incomplete result coverage: \textsc{GPT-5.6-Sol} has 117 valid questions, whereas \textsc{Claude-Opus-5}, \textsc{GLM-5.3}, \textsc{Qwen3.8-27B}, and \textsc{Ling-3.0-Flash-Fin} each have 122. Missing outputs remain in the fixed 123-question denominator, with all associated rubric items marked as failed.
}
\label{tab:overall-results}
\scriptsize
\setlength{\tabcolsep}{2.8pt}
\resizebox{\textwidth}{!}{%
\begin{tabular}{@{}lccccccccc@{}}
\toprule
\multirow{2}{*}{\textbf{Model}}
& \multicolumn{3}{c}{\textbf{Overall metrics (\%)}}
& \multicolumn{3}{c}{\textbf{Rubric dimensions (\%)}}
& \multicolumn{3}{c}{\textbf{Tool efficiency}} \\
\cmidrule(lr){2-4}\cmidrule(lr){5-7}\cmidrule(lr){8-10}
& {\fontsize{6.5pt}{7pt}\selectfont\bfseries Atomic}
& {\fontsize{6.5pt}{7pt}\selectfont\bfseries Loose Pass}
& {\fontsize{6.5pt}{7pt}\selectfont\bfseries Strict Pass}
& {\fontsize{6.5pt}{7pt}\selectfont\bfseries\shortstack{Raw-\\information}}
& {\fontsize{6.5pt}{7pt}\selectfont\bfseries\shortstack{Source\\verification}}
& {\fontsize{6.5pt}{7pt}\selectfont\bfseries\shortstack{Computation\\\& answer}}
& {\fontsize{6.5pt}{7pt}\selectfont\bfseries\shortstack{Avg.\\rounds $\downarrow$}}
& {\fontsize{6.5pt}{7pt}\selectfont\bfseries\shortstack{Avg.\\ tool calls $\downarrow$}}
& {\fontsize{6.5pt}{7pt}\selectfont\bfseries\shortstack{Max.\\ tool calls $\downarrow$}} \\
\midrule
\textsc{GPT-5.6-Sol}$^{\dagger}$            & 85.45          & 85.92          & \textbf{71.54} & 86.85          & 88.68          & 81.58          & 24.64 & 62.28 & 281 \\
\textsc{Claude-Opus-5}$^{\dagger}$          & \textbf{87.59} & \textbf{87.61} & 69.11          & \textbf{88.98} & \textbf{89.59} & \textbf{82.72} & 14.80 & 23.89 & 153 \\
\textsc{Gemini-3.7-Flash}                    & 75.89          & 76.75          & 44.72          & 81.19          & 66.62          & 72.80          & 23.73 & 23.73 &  74 \\
\textsc{Qwen3.8-Max}                         & 78.17          & 77.40          & 54.47          & 80.33          & 77.75          & 69.72          & 32.20 & 41.47 & 167 \\
\textsc{Qwen3.8-Flash}                       & 82.31          & 81.23          & 61.79          & 84.29          & 83.36          & 71.93          & 25.00 & 35.37 & 134 \\
\noalign{\vskip 0.08cm}
\hdashline
\noalign{\vskip 0.08cm}
\textsc{DeepSeek-V4-Pro}                & 76.03          & 75.41          & 51.22          & 80.09          & 73.57          & 64.88          & 21.78 & 22.21 & 100 \\

\textsc{GLM-5.3}$^{\dagger}$                & 79.32          & 80.61          & 60.98          & 83.11          & 78.56          & 75.77          & 18.61 & 24.11 &  95 \\
\textsc{GLM-5.3-Flash}                       & 79.60          & 76.64          & 56.91          & 79.50          & 80.77          & 66.44          & 22.43 & 30.46 & 135 \\
\textsc{GLM-5.2}                             & 70.61          & 65.89          & 43.09          & 69.75          & 68.30          & 54.37          & 20.34 & 23.72 &  87 \\
\textsc{Kimi-K3}                             & 84.45          & 80.83          & 59.35          & 84.84          & 70.70          & 77.98          & 17.59 & 36.49 & 313 \\
\textsc{DeepSeek-V4-Flash}              & 72.90          & 71.37          & 48.78          & 77.49          & 71.65          & 55.69          & 16.49 & 21.96 & 106 \\
\textsc{Qwen3.8-27B}$^{\dagger}$            & 78.03          & 77.28          & 55.28          & 81.49          & 76.83          & 66.92          & 28.20 & 37.33 & 149 \\
\textsc{MiniMax-M3}                          & 63.05          & 60.70          & 37.40          & 65.51          & 62.97          & 46.87          & 27.45 & 42.28 & 177 \\
\textsc{Hunyuan3-Thinking}                   & 65.76          & 65.22          & 40.65          & 70.43          & 68.01          & 50.02          & 29.24 & 40.28 & 189 \\
\textsc{Ling-3.0-Flash-Fin}$^{\dagger}$     & 75.89          & 75.07          & 52.85          & 78.43          & 82.45          & 61.25          & 23.08 & 30.87 & 119 \\
\bottomrule
\end{tabular}%
}
\end{table*}

\subsection{Unsupported-Correct Answers}

Final-answer accuracy alone cannot distinguish a well-supported result from one reached without a complete evidence chain. To quantify this distinction, we partition each decomposable question into an answer layer and an evidence layer: computation-and-answer criteria form the former when available, while raw-information criteria serve this role for direct-retrieval questions, and all remaining criteria constitute the latter. An answer is deemed correct only when all answer-layer criteria pass, and fully supported only when every evidence criterion additionally passes. This yields four outcomes: \textbf{Q1}, incorrect with no credited support; \textbf{Q2}, incorrect with partial research progress; \textbf{Q3}, correct but incompletely supported; and \textbf{Q4}, correct and fully traceable.

For model $m$, we define the \textbf{Unsupported-Correct Rate} as
\begin{equation}
\mathrm{UCR}_m =
\frac{N_m(Q3)}{N_m(Q3)+N_m(Q4)},
\end{equation}

where the denominator contains all correct-answer attempts. UCR directly captures the value of the evidence-aware atomic rubrics in \benchmarkname{}: a lower value indicates that correct answers are more consistently accompanied by complete, verifiable support. Across the 1,770 decomposable attempts, 1,150 answers are correct, of which 201 fall into Q3, yielding a micro-averaged UCR of 17.48\% and an unweighted mean across models of 17.74\%---meaning that final-answer-only evaluation would overstate fully traceable success for nearly one in six correct-answer attempts. \textsc{GPT-5.6-Sol} achieves the lowest UCR (8.70\%) and \textsc{Gemini-3.7-Flash} the highest (34.62\%). \textsc{Ling-3.0-Flash-Fin} reaches 10.00\%---third-lowest overall and below every evaluated open-weight model---indicating that its correct answers are typically backed by complete evidence chains. 

\begin{table*}[t]
\centering
\caption{Answer correctness and evidence completeness. 
NA denotes questions without independently separable answer and evidence layers. UCR is $Q3/(Q3+Q4)$, with $\downarrow$ indicating that lower is better.
}
\label{tab:unsupported-correct}
\scriptsize
\setlength{\tabcolsep}{4.6pt}
\begin{tabular}{@{}lcccccc@{}}
\toprule
\textbf{Model} & \textbf{\shortstack{Q1: Wrong,\\no support}} & \textbf{\shortstack{Q2: Wrong,\\partial process}} & \textbf{\shortstack{Q3: Correct,\\partial evidence}} & \textbf{\shortstack{Q4: Correct,\\fully traceable}} & \textbf{NA} & \textbf{UCR (\%) $\downarrow$} \\
\midrule
\textsc{GPT-5.6-Sol}$^{\dagger}$        &  7 & 19 &  8 & 84 & 5 & \textbf{8.70} \\
\textsc{Claude-Opus-5}$^{\dagger}$      &  1 & 26 &  9 & 82 & 5 &  9.89 \\
\textsc{Gemini-3.7-Flash}                &  9 & 31 & 27 & 51 & 5 & 34.62 \\
\textsc{Qwen3.8-Max}                     &  8 & 30 & 17 & 63 & 5 & 21.25 \\
\textsc{Qwen3.8-Flash}                   &  8 & 26 & 11 & 73 & 5 & 13.10 \\
\noalign{\vskip 0.08cm}
\hdashline
\noalign{\vskip 0.08cm}
\textsc{DeepSeek-V4-Pro}                 & 10 & 36 & 13 & 59 & 5 & 18.06 \\

\textsc{GLM-5.3}$^{\dagger}$            &  9 & 25 & 14 & 70 & 5 & 16.67 \\
\textsc{GLM-5.3-Flash}                   & 10 & 31 &  9 & 68 & 5 & 11.69 \\
\textsc{GLM-5.2}                         & 17 & 36 & 14 & 51 & 5 & 21.54 \\
\textsc{Kimi-K3}                         & 10 & 19 & 19 & 70 & 5 & 21.35 \\
\textsc{DeepSeek-V4-Flash}               & 13 & 35 & 12 & 58 & 5 & 17.14 \\
\textsc{Qwen3.8-27B}$^{\dagger}$        &  7 & 29 & 17 & 65 & 5 & 20.73 \\
\textsc{MiniMax-M3}                      & 20 & 41 & 12 & 45 & 5 & 21.05 \\
\textsc{Hunyuan3-Thinking}               & 17 & 42 & 12 & 47 & 5 & 20.34 \\
\textsc{Ling-3.0-Flash-Fin}$^{\dagger}$ &  7 & 41 &  7 & 63 & 5 & 10.00 \\
\midrule
\textbf{Average} & 10.20 & 31.13 & 13.40 & 63.27 & 5.00 & 17.74 \\
\bottomrule
\end{tabular}
\end{table*}

\subsection{Slice Analysis}

\begin{figure*}[t]
\centering
\includegraphics[width=\textwidth]{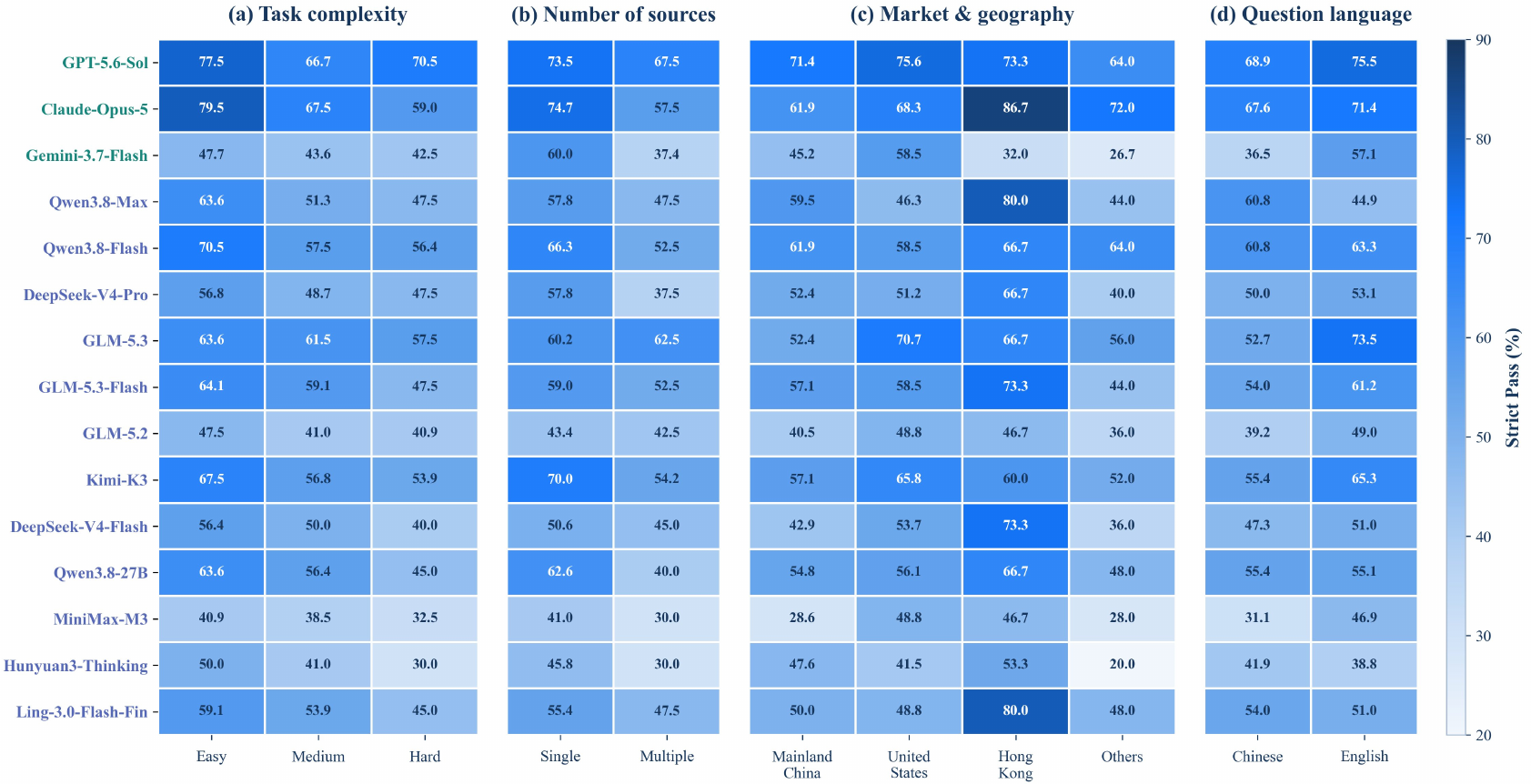}
\caption{Per-model Strict Pass across task complexity, source count, market and geography, and language. Model names are shown in \textcolor[HTML]{16877C}{\textbf{Green}} for U.S.-developed models and \textcolor[HTML]{5668B8}{\textbf{Blue}} for China-developed models.}
\label{fig:slice-strict-summary}
\end{figure*}

Figure~\ref{fig:slice-strict-summary} reports per-model Strict Pass across four dataset dimensions, with missing outputs counted as failures. Task complexity yields the clearest trend: cross-model averages decline from 60.55\% on easy questions to 52.90\% on medium and 47.71\% on hard, with 14 of 15 models following this monotonic decline. Source multiplicity produces a comparably consistent effect, as single-source questions average 58.54\% versus 46.94\% for multi-source questions, with only \textsc{GLM-5.3} reversing the pattern. Market-level performance is more heterogeneous: Hong Kong questions record the highest average (64.81\%), followed by the United States (56.74\%), mainland China (52.22\%), and other markets (45.25\%). Within this dimension, U.S.-developed models hold a clear advantage on U.S. questions (67.47\% vs. 54.06\% for China-developed models), whereas China-developed models show a narrow edge on Hong Kong questions (65.01\% vs. 64.00\%), suggesting partial regional specialization rather than uniform home-market dominance. Finally, English questions outscore Chinese questions on average (57.14\% vs. 51.71\%), though model-level exceptions---notably \textsc{Qwen3.8-Max} and \textsc{Ling-3.0-Flash-Fin}---indicate that this aggregate gap partly reflects differences in market and task composition rather than language alone.

\section{Related Work}
\label{sec:related}

\subsection{Financial QA Benchmarks}
Early financial benchmarks primarily evaluate language understanding and numerical reasoning within a bounded evidence setting. FinQA \citep{finqa} and ConvFinQA \citep{convfinqa} pair questions with pre-selected passages and tables from annual reports, and additionally annotate executable reasoning programs or conversational reasoning chains. 
TAT-QA \citep{tatqa} focuses on arithmetic reasoning over supplied tabular and textual evidence, while FinanceBench \citep{financebench} evaluates retrieval and question answering over a fixed corpus of public filings. 
Broader suites extend coverage to sentiment analysis, text classification, forecasting, multilinguality, and multimodality: FLUE \citep{flue}, PIXIU/FLARE \citep{pixiu}, FinBen \citep{finben}, MultiFinBen \citep{multifinben}, and CFBenchmark \citep{cfbenchmark} aggregate diverse financial NLP tasks from multiple sources.
FinEval \citep{fineval} additionally covers Chinese financial knowledge, security, and agent-oriented questions; CPA-KQA \citep{cpaqka} introduces skill-level cognitive diagnosis using professional accounting examinations; and FinDABench \citep{findabench} targets financial data analysis and visualization. 
To improve professional realism, FinanceQA \citep{financeqa} employs tasks written and verified by domain practitioners, while BizFinBench \citep{bizfinbench} grounds its Chinese tasks in queries drawn from a real-world financial application.

Despite their diversity, these benchmarks share a fundamental limitation: they assume that relevant evidence has already been provided or bounded, and thus cannot measure end-to-end performance spanning the full pipeline—from discovering and validating sources to extracting structured information and generating the final answer.

\subsection{Agentic Financial Benchmarks}
Recent benchmarks have increasingly shifted toward evaluating financial agents that retrieve and synthesize external information. 
Finance Agent Benchmark \citep{financeagent} comprises 537 expert-authored questions spanning nine research categories, and provides a model-agnostic evaluation harness with access to web search and SEC EDGAR; its atomic rubrics represent an improvement over exact-match scoring, though the published judge assesses answer content rather than the quality of the underlying retrieval or reasoning process. 
FinSearchComp \citep{finsearchcomp} contains 635 questions constructed and quality-controlled by 70 finance professionals, covering time-sensitive data fetching, historical lookup, and complex investigation tasks, yet assigns each response a binary final-answer score.
FrontierFinance \citep{frontierfinance} advances long-form investment research evaluation through 220 timestamped expert queries and 11,543 source-attributed criteria across six investor workflows, while also comparing systems built on different tool harnesses; however, its grading remains focused primarily on the content of final responses.
A separate benchmark under the same name \citep{frontierfinancecomputer} extends this line to long-horizon computer-use tasks that produce financial models and client-ready artifacts.
FinDeepIndicator \citep{findeepindicator} takes a more diagnostic approach by independently scoring formula specification, data collection, indicator calculation, and answer generation, but its scope is confined to template-derived indicator-construction questions.

Taken together, existing benchmarks in this line share three limitations: they either emphasize final-answer content without systematically diagnosing the full evidence-to-answer chain, conflate model capability with the effects of heterogeneous tool environments, or offer process-level evaluation only within a narrow task family. None simultaneously combines broad, demand-informed financial search coverage with provenance-aware grading, stage-level process diagnosis, and a common configurable evaluation harness.

\subsection{General Web Search Benchmarks}
Domain-general benchmarks provide complementary views of browsing and tool use across diverse task types. GAIA \citep{gaia} tests browsing, multimodality, reasoning, and tools on conceptually simple but real-world questions. 
BrowseComp \citep{browsecomp} and BrowseComp-ZH \citep{browsecompzh} stress multi-step web search for short, time-invariant answers, while BrowseComp-Plus \citep{browsecompplus} adds a fixed human-verified evidence corpus, controlled retrievers, and citation-retrieval metrics. WebWalkerQA \citep{webwalker} evaluates traversal through website hierarchies, and AssistantBench \citep{assistantbench} evaluates realistic and time-consuming open-web tasks. SimpleQA \citep{simpleqa} and Humanity's Last Exam \citep{hle} test short-form factuality and expert-level knowledge, respectively, without treating end-to-end web research as their primary object of evaluation. Mind2Web~2 \citep{mind2web2} uses tree-structured rubrics and judge agents to assess complex, time-varying answers and source attribution; however, it evaluates final retrieved information rather than intermediate interactions and avoids tasks that explicitly require complex reasoning or external computation. 

The central limitation of these benchmarks for our setting is their lack of financial specialization: they do not encode authoritative financial source hierarchies, reporting-period and definition alignment, strict numerical accuracy, or the evidence standards needed for auditable investment research.

\section{Conclusion}


We introduced \benchmarkname{}, an expert-authored benchmark designed to evaluate financial search agents beyond final-answer matching. Its 123 tasks are grounded in real-world financial scenarios and constructed through an 18-field taxonomy, a 6-axis coverage blueprint, a registry of 138 financial sources, contributions from over 50 finance experts, and a 6-stage quality-control pipeline. Each task includes an evidence-grounded reference package and atomic rubrics, enabling fine-grained assessment of both the answer and the research process supporting it.

Our evaluation of 15 model configurations under a unified tool environment reveals that current agents remain far from consistently completing financial search tasks. \textsc{Claude-Opus-5} achieves the highest Atomic and Loose Pass scores, at 87.59\% and 87.61\%, respectively, yet its Strict Pass rate falls to only 69.11\%.
Across models, computation and answer formation consistently lags behind raw-information acquisition, and 17.48\% of correct-answer attempts lack a complete, verifiable evidence chain. 
These findings demonstrate that final-answer accuracy alone can obscure incomplete evidence, unresolved requirements, and distinct capability bottlenecks. By making the underlying research process measurable and diagnosable, \benchmarkname{} provides a foundation for developing financial agents whose answers are not only correct, but also well-supported, traceable, and independently verifiable.

\bibliographystyle{assets/plainnat}
\bibliography{main}

\begin{thebibliography}{47}
\providecommand{\natexlab}[1]{#1}
\providecommand{\url}[1]{\texttt{#1}}
\expandafter\ifx\csname urlstyle\endcsname\relax
  \providecommand{\doi}[1]{doi: #1}\else
  \providecommand{\doi}{doi: \begingroup \urlstyle{rm}\Url}\fi

\bibitem[{Anthropic}(2026)]{claudeopus5}
{Anthropic}.
\newblock {Claude Opus 5} system card.
\newblock Official system card, 2026.
\newblock \url{https://www.anthropic.com/claude-opus-5-system-card}.

\bibitem[Bigeard et~al.(2025)Bigeard, Nashold, Krishnan, and Wu]{financeagent}
Antoine Bigeard, Langston Nashold, Rayan Krishnan, and Shirley Wu.
\newblock {Finance Agent Benchmark: Benchmarking LLMs on Real-world Financial Research Tasks}, 2025.
\newblock \url{https://arxiv.org/abs/2508.00828}.

\bibitem[Chen et~al.(2021)Chen, Chen, Smiley, Shah, Borova, Langdon, Moussa, Beane, Huang, Routledge, and Wang]{finqa}
Zhiyu Chen, Wenhu Chen, Charese Smiley, Sameena Shah, Iana Borova, Dylan Langdon, Reema Moussa, Matt Beane, Ting-Hao Huang, Bryan Routledge, and William~Yang Wang.
\newblock {F}in{QA}: A dataset of numerical reasoning over financial data.
\newblock In Marie-Francine Moens, Xuanjing Huang, Lucia Specia, and Scott Wen-tau Yih, editors, \emph{Proceedings of the 2021 Conference on Empirical Methods in Natural Language Processing}, pages 3697--3711, Online and Punta Cana, Dominican Republic, November 2021. Association for Computational Linguistics.
\newblock \doi{10.18653/v1/2021.emnlp-main.300}.
\newblock \url{https://aclanthology.org/2021.emnlp-main.300/}.

\bibitem[Chen et~al.(2022)Chen, Li, Smiley, Ma, Shah, and Wang]{convfinqa}
Zhiyu Chen, Shiyang Li, Charese Smiley, Zhiqiang Ma, Sameena Shah, and William~Yang Wang.
\newblock {C}onv{F}in{QA}: Exploring the chain of numerical reasoning in conversational finance question answering.
\newblock In Yoav Goldberg, Zornitsa Kozareva, and Yue Zhang, editors, \emph{Proceedings of the 2022 Conference on Empirical Methods in Natural Language Processing}, pages 6279--6292, Abu Dhabi, United Arab Emirates, December 2022. Association for Computational Linguistics.
\newblock \doi{10.18653/v1/2022.emnlp-main.421}.
\newblock \url{https://aclanthology.org/2022.emnlp-main.421/}.

\bibitem[Chen et~al.(2025)Chen, Ma, Zhuang, Nie, Zou, Liu, Green, Patel, Meng, Su, Sharifymoghaddam, Li, Hong, Shi, Liu, Thakur, Zhang, Gao, Chen, and Lin]{browsecompplus}
Zijian Chen, Xueguang Ma, Shengyao Zhuang, Ping Nie, Kai Zou, Andrew Liu, Joshua Green, Kshama Patel, Ruoxi Meng, Mingyi Su, Sahel Sharifymoghaddam, Yanxi Li, Haoran Hong, Xinyu Shi, Xuye Liu, Nandan Thakur, Crystina Zhang, Luyu Gao, Wenhu Chen, and Jimmy Lin.
\newblock {BrowseComp-Plus: A More Fair and Transparent Evaluation Benchmark of Deep-Research Agent}, 2025.
\newblock \url{https://arxiv.org/abs/2508.06600}.

\bibitem[Cohen(1960)]{cohen1960coefficient}
Jacob Cohen.
\newblock A coefficient of agreement for nominal scales.
\newblock \emph{Educational and psychological measurement}, 20\penalty0 (1):\penalty0 37--46, 1960.
\newblock \url{https://www.semanticscholar.org/paper/A-Coefficient-of-Agreement-for-Nominal-Scales-Cohen/9e463eefadbcd336c69270a299666e4104d50159}.

\bibitem[{DeepMind}(2026)]{gemini37flash}
{DeepMind}.
\newblock {Gemini 3.7 Flash} model card.
\newblock Official model card, August 2026.
\newblock \url{https://deepmind.google/models/model-cards/gemini-3-7-flash/}.

\bibitem[{DeepSeek-AI}(2026{\natexlab{a}})]{deepseekv4}
{DeepSeek-AI}.
\newblock {DeepSeek V4} technical documentation and model card.
\newblock Official technical documentation, April 2026{\natexlab{a}}.
\newblock \url{https://fe-static.deepseek.com/chat/transparency/deepseek-V4-model-card-EN.pdf}.

\bibitem[{DeepSeek-AI}(2026{\natexlab{b}})]{deepseekv4flash0731}
{DeepSeek-AI}.
\newblock {DeepSeek-V4-Flash-0731} model card.
\newblock Hugging Face model card, July 2026{\natexlab{b}}.
\newblock \url{https://huggingface.co/deepseek-ai/DeepSeek-V4-Flash-0731}.

\bibitem[{GLM-5 Team}(2026)]{glm5}
{GLM-5 Team}.
\newblock {GLM-5}: From vibe coding to agentic engineering.
\newblock \emph{arXiv preprint arXiv:2602.15763}, 2026.
\newblock \url{https://arxiv.org/abs/2602.15763}.

\bibitem[Gou et~al.(2025)Gou, Huang, Ning, Gu, Lin, Qi, Kopanev, Yu, Gutiérrez, Shu, Song, Wu, Chen, Moussa, Zhang, Xie, Li, Xue, Liao, Zhang, Zheng, Cai, Rozgic, Ziyadi, Sun, and Su]{mind2web2}
Boyu Gou, Zanming Huang, Yuting Ning, Yu~Gu, Michael Lin, Weijian Qi, Andrei Kopanev, Botao Yu, Bernal~Jiménez Gutiérrez, Yiheng Shu, Chan~Hee Song, Jiaman Wu, Shijie Chen, Hanane~Nour Moussa, Tianshu Zhang, Jian Xie, Yifei Li, Tianci Xue, Zeyi Liao, Kai Zhang, Boyuan Zheng, Zhaowei Cai, Viktor Rozgic, Morteza Ziyadi, Huan Sun, and Yu~Su.
\newblock {Mind2Web 2: Evaluating Agentic Search with Agent-as-a-Judge}, 2025.
\newblock \url{https://arxiv.org/abs/2506.21506}.

\bibitem[Guo et~al.(2025)Guo, Xia, Liu, Cao, Yang, Liu, Wang, Niu, Wang, Wang, Liang, Huang, Zhu, Wei, Chen, Shen, and Zhang]{fineval}
Xin Guo, Haotian Xia, Zhaowei Liu, Hanyang Cao, Zhi Yang, Zhiqiang Liu, Sizhe Wang, Jinyi Niu, Chuqi Wang, Yanhui Wang, Xiaolong Liang, Xiaoming Huang, Bing Zhu, Zhongyu Wei, Yun Chen, Weining Shen, and Liwen Zhang.
\newblock {F}in{E}val: A {C}hinese financial domain knowledge evaluation benchmark for large language models.
\newblock In Luis Chiruzzo, Alan Ritter, and Lu~Wang, editors, \emph{Proceedings of the 2025 Conference of the Nations of the Americas Chapter of the Association for Computational Linguistics: Human Language Technologies (Volume 1: Long Papers)}, pages 6258--6292, Albuquerque, New Mexico, April 2025. Association for Computational Linguistics.
\newblock ISBN 979-8-89176-189-6.
\newblock \doi{10.18653/v1/2025.naacl-long.318}.
\newblock \url{https://aclanthology.org/2025.naacl-long.318/}.

\bibitem[Hu et~al.(2025)Hu, Jiao, Liu, Ren, Wen, Zhang, Zhang, Gao, He, Hu, Liao, Wang, Yang, Yang, Yin, Zeng, Zhang, Zhang, Zhao, Zhu, Namkoong, Huang, and Tang]{finsearchcomp}
Liang Hu, Jianpeng Jiao, Jiashuo Liu, Yanle Ren, Zhoufutu Wen, Kaiyuan Zhang, Xuanliang Zhang, Xiang Gao, Tianci He, Fei Hu, Yali Liao, Zaiyuan Wang, Chenghao Yang, Qianyu Yang, Mingren Yin, Zhiyuan Zeng, Ge~Zhang, Xinyi Zhang, Xiying Zhao, Zhenwei Zhu, Hongseok Namkoong, Wenhao Huang, and Yuwen Tang.
\newblock {FinSearchComp: Towards a Realistic, Expert-Level Evaluation of Financial Search and Reasoning}, 2025.
\newblock \url{https://arxiv.org/abs/2509.13160}.

\bibitem[{Hunyuan Team}(2026)]{hunyuan3}
{Hunyuan Team}.
\newblock {Hy3}: A reasoning and agent model.
\newblock Official model repository and model card, 2026.
\newblock \url{https://github.com/Tencent-Hunyuan/Hy3}.

\bibitem[{inclusionAI}(2026)]{ling30flash-fin}
{inclusionAI}.
\newblock {Ling-3.0-Flash-Fin} model card.
\newblock Openrouter model card, 2026.
\newblock \url{https://openrouter.ai/inclusionai/ling-3.0-flash-fin:free}.

\bibitem[Islam et~al.(2023)Islam, Kannappan, Kiela, Qian, Scherrer, and Vidgen]{financebench}
Pranab Islam, Anand Kannappan, Douwe Kiela, Rebecca Qian, Nino Scherrer, and Bertie Vidgen.
\newblock {FinanceBench: A New Benchmark for Financial Question Answering}, 2023.
\newblock \url{https://arxiv.org/abs/2311.11944}.

\bibitem[{Kimi Team}(2026)]{kimik3}
{Kimi Team}.
\newblock {Kimi K3}: Open frontier intelligence.
\newblock \emph{arXiv preprint arXiv:2607.24653}, 2026.
\newblock \url{https://arxiv.org/abs/2607.24653}.

\bibitem[Krumdick et~al.(2026)Krumdick, Reddy, Chaudhary, Day, Ahmed, Haqqi, Fahim, Amjad, Orakzai, Gul, and Tanner]{frontierfinancecomputer}
Michael Krumdick, Varshini Reddy, Shivani Chaudhary, William Day, Maarij Ahmed, Hayan Haqqi, Muhammad~Ahsen Fahim, Hanzallah Amjad, Ahmad Orakzai, Aqsa Gul, and Chris Tanner.
\newblock {FrontierFinance: A Long-Horizon Computer-Use Benchmark of Real-World Financial Tasks}, 2026.
\newblock \url{https://arxiv.org/abs/2604.05912}.

\bibitem[Kuang et~al.(2025)Kuang, Zhu, Jiang, Lai, Wang, Wang, Qiu, Huang, Peng, Xie, and Ananiadou]{cpaqka}
Ziyan Kuang, Feiyu Zhu, Maowei Jiang, Yanzhao Lai, Zelin Wang, Zhitong Wang, Meikang Qiu, Jiajia Huang, Min Peng, Qianqian Xie, and Sophia Ananiadou.
\newblock {From Scores to Skills: A Cognitive Diagnosis Framework for Evaluating Financial Large Language Models}, 2025.
\newblock \url{https://arxiv.org/abs/2508.13491}.

\bibitem[Lai et~al.(2026)Lai, Xu, Yang, Chen, Xu, Zeng, Li, Sun, Zhu, Zhang, and Zhao]{minimaxmsa}
Xunhao Lai, Weiqi Xu, Yufeng Yang, Qiaorui Chen, Yang Xu, Lunbin Zeng, Xiaolong Li, Haohai Sun, Haichao Zhu, Vito Zhang, and Pengyu Zhao.
\newblock {MiniMax Sparse Attention}.
\newblock \emph{arXiv preprint arXiv:2606.13392}, 2026.
\newblock \url{https://arxiv.org/abs/2606.13392}.

\bibitem[Lei et~al.(2023)Lei, Li, Cheng, Ding, and Jiang]{cfbenchmark}
Yang Lei, Jiangtong Li, Dawei Cheng, Zhijun Ding, and Changjun Jiang.
\newblock {CFBenchmark: Chinese Financial Assistant Benchmark for Large Language Model}, 2023.
\newblock \url{https://arxiv.org/abs/2311.05812}.

\bibitem[Liu et~al.(2025)Liu, Zhao, Jia, Zhuang, Long, Zhou, Zhou, Lan, and Chong]{findabench}
Shu Liu, Shangqing Zhao, Chenghao Jia, Xinlin Zhuang, Zhaoguang Long, Jie Zhou, Aimin Zhou, Man Lan, and Yang Chong.
\newblock {F}in{DAB}ench: Benchmarking financial data analysis ability of large language models.
\newblock In Owen Rambow, Leo Wanner, Marianna Apidianaki, Hend Al-Khalifa, Barbara~Di Eugenio, and Steven Schockaert, editors, \emph{Proceedings of the 31st International Conference on Computational Linguistics}, pages 710--725, Abu Dhabi, UAE, January 2025. Association for Computational Linguistics.
\newblock \url{https://aclanthology.org/2025.coling-main.48/}.

\bibitem[Lu et~al.(2025)Lu, Guo, Zhang, Zhu, and Liu]{bizfinbench}
Guilong Lu, Xuntao Guo, Rongjunchen Zhang, Wenqiao Zhu, and Ji~Liu.
\newblock {BizFinBench: A Business-Driven Real-World Financial Benchmark for Evaluating LLMs}, 2025.
\newblock \url{https://arxiv.org/abs/2505.19457}.

\bibitem[Mateega et~al.(2025)Mateega, Georgescu, and Tang]{financeqa}
Spencer Mateega, Carlos Georgescu, and Danny Tang.
\newblock {FinanceQA: A Benchmark for Evaluating Financial Analysis Capabilities of Large Language Models}, 2025.
\newblock \url{https://arxiv.org/abs/2501.18062}.

\bibitem[Mialon et~al.(2023)Mialon, Fourrier, Swift, Wolf, LeCun, and Scialom]{gaia}
Grégoire Mialon, Clémentine Fourrier, Craig Swift, Thomas Wolf, Yann LeCun, and Thomas Scialom.
\newblock {GAIA: a benchmark for General AI Assistants}, 2023.
\newblock \url{https://arxiv.org/abs/2311.12983}.

\bibitem[{MiniMax}(2026)]{minimaxm3}
{MiniMax}.
\newblock {MiniMax M3}: Frontier coding, 1m context, native multimodality---all in one model.
\newblock Official research release, June 2026.
\newblock \url{https://www.minimax.io/blog/minimax-m3}.

\bibitem[{OpenAI}(2026)]{gpt56}
{OpenAI}.
\newblock {GPT-5.6} system card.
\newblock OpenAI Deployment Safety Hub, 2026.
\newblock \url{https://deploymentsafety.openai.com/gpt-5-6}.

\bibitem[Peng et~al.(2025)Peng, Qian, Wang, Xiang, He, Ren, Jiang, Zhang, Guo, Zhao, He, Han, Feng, Jiang, Cao, Li, Yu, Wang, Gao, Lin, Wang, Yang, Zhao, Liu, Lu, Huang, Wang, Papadopoulos, Giannouris, Soufleri, Chen, Deng, Fu, Zhao, Lin, Qiu, Smith, Cohan, Liu, Huang, Xiong, Lopez-Lira, Chen, Tsujii, Nie, Ananiadou, and Xie]{multifinben}
Xueqing Peng, Lingfei Qian, Yan Wang, Ruoyu Xiang, Yueru He, Yang Ren, Mingyang Jiang, Vincent~Jim Zhang, Yuqing Guo, Jeff Zhao, Huan He, Yi~Han, Yun Feng, Yuechen Jiang, Yupeng Cao, Haohang Li, Yangyang Yu, Xiaoyu Wang, Penglei Gao, Shengyuan Lin, Keyi Wang, Shanshan Yang, Yilun Zhao, Zhiwei Liu, Peng Lu, Jerry Huang, Suyuchen Wang, Triantafillos Papadopoulos, Polydoros Giannouris, Efstathia Soufleri, Nuo Chen, Zhiyang Deng, Heming Fu, Yijia Zhao, Mingquan Lin, Meikang Qiu, Kaleb~E Smith, Arman Cohan, Xiao-Yang Liu, Jimin Huang, Guojun Xiong, Alejandro Lopez-Lira, Xi~Chen, Junichi Tsujii, Jian-Yun Nie, Sophia Ananiadou, and Qianqian Xie.
\newblock {MultiFinBen: Benchmarking Large Language Models for Multilingual and Multimodal Financial Application}, 2025.
\newblock \url{https://arxiv.org/abs/2506.14028}.

\bibitem[Phan et~al.(2025)]{hle}
Long Phan et~al.
\newblock {Humanity's Last Exam}, 2025.
\newblock \url{https://arxiv.org/abs/2501.14249}.

\bibitem[{Qwen Team}(2026{\natexlab{a}})]{qwen3827b}
{Qwen Team}.
\newblock {Qwen3.8-27B} model card.
\newblock Hugging Face model card, August 2026{\natexlab{a}}.
\newblock \url{https://huggingface.co/Qwen/Qwen3.8-27B}.

\bibitem[{Qwen Team}(2026{\natexlab{b}})]{qwen38flash}
{Qwen Team}.
\newblock {Qwen3.8-Flash} model documentation.
\newblock Qwen Cloud documentation, August 2026{\natexlab{b}}.
\newblock \url{https://docs.qwencloud.com/developer-guides/getting-started/vision-models}.

\bibitem[{Qwen Team}(2026{\natexlab{c}})]{qwen38max}
{Qwen Team}.
\newblock {Qwen3.8-Max}: A new bar for coding and cowork.
\newblock Official model release, August 2026{\natexlab{c}}.
\newblock \url{https://qwen.ai/blog?id=qwen3.8}.

\bibitem[Shah et~al.(2022)Shah, Chawla, Eidnani, Shah, Du, Chava, Raman, Smiley, Chen, and Yang]{flue}
Raj Shah, Kunal Chawla, Dheeraj Eidnani, Agam Shah, Wendi Du, Sudheer Chava, Natraj Raman, Charese Smiley, Jiaao Chen, and Diyi Yang.
\newblock When {FLUE} meets {FLANG}: Benchmarks and large pretrained language model for financial domain.
\newblock In Yoav Goldberg, Zornitsa Kozareva, and Yue Zhang, editors, \emph{Proceedings of the 2022 Conference on Empirical Methods in Natural Language Processing}, pages 2322--2335, Abu Dhabi, United Arab Emirates, December 2022. Association for Computational Linguistics.
\newblock \doi{10.18653/v1/2022.emnlp-main.148}.
\newblock \url{https://aclanthology.org/2022.emnlp-main.148/}.

\bibitem[Wei et~al.(2024)Wei, Karina, Chung, Jiao, Papay, Glaese, Schulman, and Fedus]{simpleqa}
Jason Wei, Nguyen Karina, Hyung~Won Chung, Yunxin~Joy Jiao, Spencer Papay, Amelia Glaese, John Schulman, and William Fedus.
\newblock {Measuring short-form factuality in large language models}, 2024.
\newblock \url{https://arxiv.org/abs/2411.04368}.

\bibitem[Wei et~al.(2025)Wei, Sun, Papay, McKinney, Han, Fulford, Chung, Passos, Fedus, and Glaese]{browsecomp}
Jason Wei, Zhiqing Sun, Spencer Papay, Scott McKinney, Jeffrey Han, Isa Fulford, Hyung~Won Chung, Alex~Tachard Passos, William Fedus, and Amelia Glaese.
\newblock {BrowseComp: A Simple Yet Challenging Benchmark for Browsing Agents}, 2025.
\newblock \url{https://arxiv.org/abs/2504.12516}.

\bibitem[Wu et~al.(2025)Wu, Yin, Jiang, Wang, Xi, Fang, Zhang, He, Zhou, Xie, and Huang]{webwalker}
Jialong Wu, Wenbiao Yin, Yong Jiang, Zhenglin Wang, Zekun Xi, Runnan Fang, Linhai Zhang, Yulan He, Deyu Zhou, Pengjun Xie, and Fei Huang.
\newblock {WebWalker: Benchmarking LLMs in Web Traversal}, 2025.
\newblock \url{https://arxiv.org/abs/2501.07572}.

\bibitem[Xie et~al.(2023)Xie, Han, Zhang, Lai, Peng, Lopez-Lira, and Huang]{pixiu}
Qianqian Xie, Weiguang Han, Xiao Zhang, Yanzhao Lai, Min Peng, Alejandro Lopez-Lira, and Jimin Huang.
\newblock {PIXIU: A Large Language Model, Instruction Data and Evaluation Benchmark for Finance}, 2023.
\newblock \url{https://arxiv.org/abs/2306.05443}.

\bibitem[Xie et~al.(2024)Xie, Han, Chen, Xiang, Zhang, He, Xiao, Li, Dai, Feng, Xu, Kang, Kuang, Yuan, Yang, Luo, Zhang, Liu, Xiong, Deng, Jiang, Yao, Li, Yu, Hu, Huang, Liu, Lopez-Lira, Wang, Lai, Wang, Peng, Ananiadou, and Huang]{finben}
Qianqian Xie, Weiguang Han, Zhengyu Chen, Ruoyu Xiang, Xiao Zhang, Yueru He, Mengxi Xiao, Dong Li, Yongfu Dai, Duanyu Feng, Yijing Xu, Haoqiang Kang, Ziyan Kuang, Chenhan Yuan, Kailai Yang, Zheheng Luo, Tianlin Zhang, Zhiwei Liu, Guojun Xiong, Zhiyang Deng, Yuechen Jiang, Zhiyuan Yao, Haohang Li, Yangyang Yu, Gang Hu, Jiajia Huang, Xiao-Yang Liu, Alejandro Lopez-Lira, Benyou Wang, Yanzhao Lai, Hao Wang, Min Peng, Sophia Ananiadou, and Jimin Huang.
\newblock {FinBen: A Holistic Financial Benchmark for Large Language Models}, 2024.
\newblock \url{https://arxiv.org/abs/2402.12659}.

\bibitem[Yang et~al.(2026)Yang, Zhu, Lin, Bai, Liu, Huang, Zimmermann, and Chua]{findeepindicator}
Chaoqun Yang, Fengbin Zhu, Xinyu Lin, Long Bai, Xiaoluan Liu, Ke-Wei Huang, Roger Zimmermann, and Tat-Seng Chua.
\newblock {FinDeepIndicator: Benchmarking Deep Research Agents in End-to-End Financial Indicator Construction}, 2026.
\newblock \url{https://arxiv.org/abs/2608.00764}.

\bibitem[Yao et~al.(2023)Yao, Zhao, Yu, Du, Shafran, Narasimhan, and Cao]{yao2023reactsynergizingreasoningacting}
Shunyu Yao, Jeffrey Zhao, Dian Yu, Nan Du, Izhak Shafran, Karthik Narasimhan, and Yuan Cao.
\newblock React: Synergizing reasoning and acting in language models, 2023.
\newblock \url{https://arxiv.org/abs/2210.03629}.

\bibitem[Yoran et~al.(2024)Yoran, Amouyal, Malaviya, Bogin, Press, and Berant]{assistantbench}
Ori Yoran, Samuel~Joseph Amouyal, Chaitanya Malaviya, Ben Bogin, Ofir Press, and Jonathan Berant.
\newblock {AssistantBench: Can Web Agents Solve Realistic and Time-Consuming Tasks?}, 2024.
\newblock \url{https://arxiv.org/abs/2407.15711}.

\bibitem[{Z.ai}(2026{\natexlab{a}})]{glm52}
{Z.ai}.
\newblock {GLM-5.2}: Built for long-horizon tasks.
\newblock Official research release, June 2026{\natexlab{a}}.
\newblock \url{https://z.ai/blog/glm-5.2}.

\bibitem[{Z.ai}(2026{\natexlab{b}})]{glm53}
{Z.ai}.
\newblock {GLM-5.3}: Frontier coding with emergent cyber capabilities.
\newblock Official research release, August 2026{\natexlab{b}}.
\newblock \url{https://z.ai/blog/glm-5.3}.

\bibitem[{Z.ai}(2026{\natexlab{c}})]{glm53flash}
{Z.ai}.
\newblock {GLM-5.3-Flash}: Frontier intelligence, flash cost.
\newblock Official research release, August 2026{\natexlab{c}}.
\newblock \url{https://z.ai/blog/glm-5.3-flash}.

\bibitem[Zhang et~al.(2026)Zhang, Koyluoglu, Venkatesh, Martinez, Bhatia, Alidoust, and Paranjape]{frontierfinance}
Yuhao Zhang, O.~Ozan Koyluoglu, Thejas Venkatesh, Richard~Diehl Martinez, Vishank Bhatia, Arash Alidoust, and Ashwin Paranjape.
\newblock {FrontierFinance: A Challenging Benchmark for Measuring Frontier Intelligence of Finance Agents}, 2026.
\newblock \url{https://arxiv.org/abs/2608.11683}.

\bibitem[Zhou et~al.(2025)Zhou, Leon, Ying, Zhang, Shao, Ye, Chong, Jin, Xie, Cao, Gu, Hong, Ren, Chen, Liu, and Hua]{browsecompzh}
Peilin Zhou, Bruce Leon, Xiang Ying, Can Zhang, Yifan Shao, Qichen Ye, Dading Chong, Zhiling Jin, Chenxuan Xie, Meng Cao, Yuxin Gu, Sixin Hong, Jing Ren, Jian Chen, Chao Liu, and Yining Hua.
\newblock {BrowseComp-ZH: Benchmarking Web Browsing Ability of Large Language Models in Chinese}, 2025.
\newblock \url{https://arxiv.org/abs/2504.19314}.

\bibitem[Zhu et~al.(2021)Zhu, Lei, Huang, Wang, Zhang, Lv, Feng, and Chua]{tatqa}
Fengbin Zhu, Wenqiang Lei, Youcheng Huang, Chao Wang, Shuo Zhang, Jiancheng Lv, Fuli Feng, and Tat-Seng Chua.
\newblock {TAT}-{QA}: A question answering benchmark on a hybrid of tabular and textual content in finance.
\newblock In Chengqing Zong, Fei Xia, Wenjie Li, and Roberto Navigli, editors, \emph{Proceedings of the 59th Annual Meeting of the Association for Computational Linguistics and the 11th International Joint Conference on Natural Language Processing (Volume 1: Long Papers)}, pages 3277--3287, Online, August 2021. Association for Computational Linguistics.
\newblock \doi{10.18653/v1/2021.acl-long.254}.
\newblock \url{https://aclanthology.org/2021.acl-long.254/}.

\end{thebibliography}

\clearpage
\appendix

\section{Tool Environment}
\label{app:tool-environment}

Our evaluation environment provides three complementary tools for information discovery, evidence inspection, and numerical analysis. All evaluated models receive the same tool descriptions, interfaces, execution limits, and per-task budgets. The reported leaderboard excludes runs that use additional professional financial tools, so differences in performance are not attributable to unequal access to proprietary databases or model-specific plugins.

\subsection{Tool Interfaces}

Table~\ref{tab:tool-environment} summarizes the common interfaces. The schemas are intentionally compact: search discovers candidate sources, visit exposes the contents of a selected source, and Python performs deterministic computation over retrieved information. Search-result snippets are treated only as navigation aids; evidence-based answers are expected to rely on content inspected through the visit tool.

\begin{table*}[t]
\centering
\caption{Tools available to every evaluated agent in the common basic-tool environment. 
}
\label{tab:tool-environment}
\small
\setlength{\tabcolsep}{5pt}
\begin{tabularx}{\textwidth}{@{}p{0.14\textwidth}p{0.22\textwidth}Xp{0.25\textwidth}@{}}
\toprule
\textbf{Tool} & \textbf{Primary input} & \textbf{Returned information} & \textbf{Role in the workflow} \\
\midrule
Web Search
& A textual search query
& A ranked list of candidate results, including the title, URL, and snippet when available
& Discover potentially relevant official filings, company disclosures, statistics, reports, databases, and other public sources \\

Visit
& One or more target URLs
& Extracted content and available document metadata from HTML webpages or PDF documents
& Inspect a selected source, locate supporting passages or tables, and retrieve evidence used in the answer \\

Python
& Executable Python code
& Standard output, standard error, and execution status from an isolated runtime with common numerical and data-analysis libraries
& Perform arithmetic, aggregation, unit conversion, filtering, tabular manipulation, and consistency checks \\
\bottomrule
\end{tabularx}
\end{table*}

\noindent\textbf{Web Search.}
The search tool accesses search results via SerpAPI\footnote{\url{https://serpapi.com/}}. It accepts a natural-language query and returns ranked candidate pages. Agents may issue multiple calls and reformulate queries as the research process develops. Each result provides sufficient routing information to decide whether the underlying page should be inspected. Because snippets can be incomplete, stale, or detached from their original context, they are not treated as authoritative evidence by themselves.

\noindent\textbf{Visit.}
The visit tool employs Jina Reader\footnote{\url{https://jina.ai/}} to retrieve and parse the contents of a specified public URL. It supports both webpages and PDF documents and returns machine-readable text together with available structural or document metadata. The tool is used to inspect primary materials, identify exact evidence locations, and verify entities, periods, units, definitions, and data versions. If extraction fails or a page is inaccessible, the failure is returned to the agent, which may search for an alternative public source within the remaining budget.

\noindent\textbf{Python.}
The Python tool executes code in an isolated environment with common numerical and data-analysis packages. It supports calculations and transformations that would be error-prone to perform manually, including multi-period aggregation, percentage and growth-rate computation, unit normalization, conditional filtering, and tabular checks. The tool does not provide independent financial evidence: all numerical inputs must still be grounded in sources retrieved through the search and visit workflow.




\subsection{Controlled Execution}

All models are evaluated through the same ReAct-style loop. At each step, an agent may produce a final response or invoke one of the three tools; the resulting observation is appended to the interaction history before the next model call. Tool-call budgets, timeouts, network conditions, and termination rules are held fixed across model configurations. Tool or model-service failures are recorded explicitly. Under the fixed-123-question protocol used in the main results, an unresolved execution or judging failure remains in the denominator and receives zero credit rather than being silently removed.



\section{Evaluated Models}
\label{app:model-cards}

Table~\ref{tab:model-cards} summarizes the model configurations used in our experiments. We report the evaluated API identifier or public checkpoint whenever it is available. Parameter counts follow official model documentation; "active" denotes the parameters activated per token for sparse mixture-of-experts models, and "--" indicates that the provider has not publicly disclosed the value. \faLock{} denotes a proprietary or hosted configuration, whereas \faUnlock{} denotes publicly available model weights.

\begin{table*}[t]
\centering
\caption{Model cards for the systems evaluated in \benchmarkname{}.}
\label{tab:model-cards}
\scriptsize
\setlength{\tabcolsep}{4pt}
\resizebox{\textwidth}{!}{%
\begin{tabular}{@{}llllll@{}}
\toprule
\textsc{Model} & \textsc{Provider} & \textsc{Checkpoints} & \textsc{Avail.} & \textsc{\#Param} & \textsc{Reference} \\
\midrule
\href{https://developers.openai.com/api/docs/models/gpt-5.6-sol}{\textsc{GPT-5.6-Sol}}
& OpenAI & \texttt{gpt-5.6-sol} & \faLock & -- & \citealp{gpt56} \\

\href{https://platform.claude.com/docs/en/models/opus-5/whats-new-opus-5}{\textsc{Claude-Opus-5}}
& Anthropic & \texttt{claude-opus-5} & \faLock & -- & \citealp{claudeopus5} \\

\href{https://deepmind.google/models/model-cards/gemini-3-7-flash/}{\textsc{Gemini-3.7-Flash}}
& Google DeepMind & \texttt{gemini-3.7-flash} & \faLock & -- & \citealp{gemini37flash} \\

\href{https://help.aliyun.com/en/model-studio/qwen3-8-max}{\textsc{Qwen3.8-Max}}
& Alibaba/Qwen & \texttt{qwen3.8-max} & \faLock & 2.4T / 95B active & \citealp{qwen38max} \\

\href{https://docs.qwencloud.com/developer-guides/getting-started/vision-models}{\textsc{Qwen3.8-Flash}}
& Alibaba/Qwen & \texttt{qwen3.8-flash} & \faLock & -- & \citealp{qwen38flash} \\
\noalign{\vskip 0.08cm}
\hdashline
\noalign{\vskip 0.08cm}
\href{https://fe-static.deepseek.com/chat/transparency/deepseek-V4-model-card-EN.pdf}{\textsc{DeepSeek-V4-Pro-0813}}
& DeepSeek-AI & \texttt{DeepSeek-V4-Pro} & \faLock & 1.6T / 49B active & \citealp{deepseekv4} \\

\href{https://huggingface.co/zai-org/GLM-5.3}{\textsc{GLM-5.3}}
& Z.ai & \texttt{zai-org/GLM-5.3} & \faUnlock & 753B / 40B active & \citealp{glm53} \\

\href{https://huggingface.co/zai-org/GLM-5.3-Flash}{\textsc{GLM-5.3-Flash}}
& Z.ai & \texttt{zai-org/GLM-5.3-Flash} & \faUnlock & 320B / 18B active & \citealp{glm53flash} \\

\href{https://huggingface.co/zai-org/GLM-5.2}{\textsc{GLM-5.2}}
& Z.ai & \texttt{zai-org/GLM-5.2} & \faUnlock & 753B / 40B active & \citealp{glm52} \\

\href{https://huggingface.co/moonshotai/Kimi-K3}{\textsc{Kimi-K3}}
& Moonshot AI & \texttt{moonshotai/Kimi-K3} & \faUnlock & 2.8T / 104B active & \citealp{kimik3} \\

\href{https://huggingface.co/deepseek-ai/DeepSeek-V4-Flash-0731}{\textsc{DeepSeek-V4-Flash-0731}}
& DeepSeek-AI & \texttt{deepseek-ai/DeepSeek-V4-Flash} & \faUnlock & 285B / 13B active & \citealp{deepseekv4flash0731} \\

\href{https://huggingface.co/Qwen/Qwen3.8-27B}{\textsc{Qwen3.8-27B}}
& Alibaba/Qwen & \texttt{Qwen/Qwen3.8-27B} & \faUnlock & 27B & \citealp{qwen3827b} \\

\href{https://huggingface.co/MiniMaxAI/MiniMax-M3}{\textsc{MiniMax-M3}}
& MiniMax & \texttt{MiniMaxAI/MiniMax-M3} & \faUnlock & 428B / 23B active & \citealp{minimaxm3,minimaxmsa} \\

\href{https://github.com/Tencent-Hunyuan/Hy3}{\textsc{Hunyuan3-Thinking}}
& Tencent & \texttt{tencent/Hy3} & \faUnlock & 295B / 21B active & \citealp{hunyuan3} \\

\href{https://openrouter.ai/inclusionai/ling-3.0-flash-fin:free}{\textsc{Ling-3.0-Flash-Fin}}
& inclusionAI & \texttt{Ling-3.0-Flash-Fin} & \faUnlock & 124B / 5.1B active & \citealp{ling30flash-fin} \\

\bottomrule
\end{tabular}%
}
\end{table*}

\section{Judge Prompt Template}
\label{app:judge-prompt}



Table~\ref{tab:judge-prompt} provides the Judge Prompt Template; Chinese queries use a corresponding Chinese version. The placeholders are populated with the evaluation date, question, atomic rubric items, and model response for each instance. Since the judge observes only the final response—not the agent's hidden reasoning or tool trajectory—source-verification and traceability scores reflect the evidence presented in the final response rather than every latent action taken during retrieval. Recorded trajectories are analyzed separately for tool-use statistics and qualitative case studies, and are not used to overwrite rubric decisions in the main leaderboard.

\definecolor{JudgePromptBlue}{HTML}{1478FF}
\definecolor{JudgePromptFrame}{HTML}{17375E}
\definecolor{JudgePromptBack}{HTML}{F3F4F6}

\clearpage
\begin{table*}[!t]
\centering
\caption{Prompt template for the automatic rubric judge.}
\label{tab:judge-prompt}

\begin{tcolorbox}[
    enhanced jigsaw,
    colback=JudgePromptBack,
    colframe=JudgePromptFrame,
    boxrule=0.7pt,
    arc=3mm,
    outer arc=3mm,
    left=4mm,
    right=4mm,
    top=3mm,
    bottom=3mm,
    before skip=8pt,
    after skip=8pt
]

\noindent{\color{JudgePromptBlue}\bfseries\large Automatic Rubric-Judge Prompt}

\medskip
\small

You are a strict and impartial judge for financial-search evaluation. Given a question and its itemized rubric, determine whether the model response satisfies each rubric item.

\medskip
\noindent\textbf{Current evaluation date:}

\noindent\texttt{\{current\_date\}}

\medskip
\noindent\textbf{Evaluation requirements:}

\begin{enumerate}
    \item Base your decisions solely on the itemized rubric. Do not refer to any reference answer or external ground truth.
    \item Evaluate every rubric item independently and determine whether it is fully satisfied. Do not output a score for any item; scores are calculated by the evaluation system.
    \item Strictly verify numerical values, units, time ranges, measurement definitions and conventions, data sources, calculation procedures, and final-answer requirements.
    \item If the model provides only a conclusion but omits a key source or calculation step required by the rubric, mark the corresponding rubric item as not satisfied.
    \item If the model uses an incorrect source, year, measurement definition, or convention, mark the corresponding rubric item as not satisfied.
    \item The \texttt{category} annotation of each rubric item is used only for subsequent attribution analysis and must not affect the strictness with which that item is evaluated.
    \item For each rubric item, output only \texttt{passed=true} or \texttt{passed=false}. Output \texttt{true} only if the item is fully satisfied; otherwise, output \texttt{false}.
    \item Treat facts, numerical values, and dates explicitly stated in the rubric as authoritative benchmark annotations. Do not reject information marked as correct in the rubric based on your own assessment of dates or data-release timing.
    \item If the rubric requires an official source, a specific source, or a specific definition, strictly verify that the model response satisfies the corresponding source and definitional requirements.
    \item The output must be valid JSON. Do not enclose it in a Markdown code block.
\end{enumerate}

\noindent\textbf{Question:}

\noindent\texttt{\{question\}}

\medskip
\noindent\textbf{Itemized rubric:}

 (JSON; item numbers and point values have been removed from the rubric text, and \texttt{category} is used only for attribution analysis)

\noindent\texttt{\{rubric\_items\}}

\medskip
\noindent\textbf{Model response:}

\noindent\texttt{\{model\_answer\}}

\medskip
\noindent\textbf{Return JSON in the following format:}

\medskip
{\ttfamily\footnotesize
\{\par
\hspace*{1em}"rubric\_item\_judgements": [\par
\hspace*{2em}\{\par
\hspace*{3em}"id": 1,\par
\hspace*{3em}"passed": true,\par
\hspace*{3em}"category": "category from the corresponding input item",\par
\hspace*{3em}"reason": "primary basis for determining whether the item is satisfied"\par
\hspace*{2em}\}\par
\hspace*{1em}],\par
\hspace*{1em}"summary": "one-sentence summary of the main strengths and deficiencies",\par
\hspace*{1em}"deductions": ["major deduction 1", "major deduction 2"]\par
\}\par
}

\noindent\textbf{Note:} Do not output \texttt{total\_score}, \texttt{score}, or \texttt{max\_score}. The evaluation system calculates the final score for each rubric item from its \texttt{passed} value.

\end{tcolorbox}
\end{table*}

\end{document}